\documentclass[11pt]{article}

\PassOptionsToPackage{dvipsnames,table}{xcolor}

\usepackage[preprint]{acl}
\ifdefined\pdfsuppressptexinfo\pdfsuppressptexinfo=-1\fi

\usepackage{times}
\usepackage{latexsym}
\usepackage[T1]{fontenc}
\usepackage[utf8]{inputenc}
\usepackage{microtype}
\usepackage{inconsolata}
\usepackage{graphicx}

\usepackage{amsmath}
\usepackage{amssymb}
\usepackage{amsfonts}
\usepackage{mathtools}
\usepackage{algorithm}
\usepackage[noend]{algpseudocode}
\usepackage[algo2e,vlined,ruled]{algorithm2e}
\usepackage{booktabs}
\usepackage{multirow}
\usepackage{makecell}
\usepackage{tabularx}
\usepackage{tabularray}
\usepackage{caption}
\usepackage{subcaption}
\usepackage{capt-of}
\usepackage{float}
\usepackage{placeins}
\usepackage{listings}
\usepackage{nicefrac}
\usepackage{enumerate}
\usepackage{enumitem}
\usepackage{xcolor}
\usepackage{colortbl}
\usepackage{pgfplots}
\usepackage{pdfpages}
\usepackage[most]{tcolorbox}
\pgfplotsset{compat=1.18}

\algrenewcommand\algorithmiccomment[1]{\hfill{\(\triangleright\) #1}}
\makeatletter
\algrenewcommand\ALG@beginalgorithmic{\small}
\makeatother

\newtcolorbox{softbluebox}{
  enhanced,
  colback=cyan!7!white,
  colframe=cyan!35!black,
  boxrule=0.4pt,
  arc=6pt,
  left=10pt,right=10pt,top=8pt,bottom=8pt,
  sharp corners=false
}

\newtcolorbox{inferencebox}{
  enhanced,
  colback=cyan!6!white,
  colframe=cyan!40!gray,
  boxrule=0.4pt,
  arc=3pt,
  left=3pt, right=3pt, top=8pt, bottom=3pt,
  before skip=2pt,
  after skip=6pt,
}

\newcommand{\RoleLib}{\mathcal{R}}
\newcommand{\RFC}{\mathrm{RFC}}
\newcommand{\GR}{\mathrm{GR}}

\definecolor{forestgreen}{HTML}{228B22}
\definecolor{mortargrey}{HTML}{595959}
\definecolor{macandcheese}{HTML}{FFBC79}
\definecolor{darkred}{HTML}{B3412C}
\definecolor{darkblue}{HTML}{326B89}
\definecolor{coralred}{HTML}{FF6666}
\definecolor{creamoderate}{HTML}{FFEB99}
\definecolor{lightgreen}{HTML}{A0D798}

\definecolor{TaskColor}{HTML}{F6F6F6}
\definecolor{RouterColor}{HTML}{FFE5B4}
\definecolor{ChainColor}{HTML}{E6EEF5}
\definecolor{ValidatorColor}{HTML}{F0FFF0}
\definecolor{AddOpColor}{HTML}{90EE90}
\definecolor{PruneOpColor}{HTML}{FFB6C1}
\definecolor{SwapOpColor}{HTML}{ADD8E6}
\definecolor{ReRouteOpColor}{HTML}{FFDAB9}
\definecolor{OursBg}{HTML}{FFF8E7}

\usepackage{pifont}
\newif\ifTOneProps \TOnePropsfalse %
\newif\ifTOneAvg   \TOneAvgtrue    %

\providecolor{OursBg}{HTML}{FFF8E7}       %
\providecolor{forestgreen}{HTML}{228B22}
\providecolor{darkred}{HTML}{B3412C}
\definecolor{phgray}{HTML}{A0A0A0}
\definecolor{TOneQA}{HTML}{DDE1F1}       %
\definecolor{TOneReason}{HTML}{EEE6F1}
\definecolor{TOneCode}{HTML}{E0EBD7}
\definecolor{TOneWeb}{HTML}{FDF2D1}
\definecolor{TOneAvg}{HTML}{EDF0F6}      %
\definecolor{TOneFam}{HTML}{F4F4F4}      %
\definecolor{TOneExt}{HTML}{E5F3EF}      %
\providecolor{OldMark}{HTML}{B06A00}     %

\providecommand{\best}{}\renewcommand{\best}[1]{\textbf{#1}}
\providecommand{\second}{}\renewcommand{\second}[1]{\underline{#1}}
\providecommand{\sd}{}\renewcommand{\sd}[1]{\textsubscript{$\pm$#1}}                    %

\providecommand{\ddown}{}\renewcommand{\ddown}[1]{\textcolor{darkred}{$-$#1}}
\providecommand{\oc}{}\renewcommand{\oc}{\cellcolor{OursBg}}
\providecommand{\ac}{}\renewcommand{\ac}{\cellcolor{TOneAvg}}
\providecommand{\pyes}{}\renewcommand{\pyes}{\ding{51}}                              %
\providecommand{\pno}{}\renewcommand{\pno}{\textcolor{black!40}{\ding{55}}}           %
\providecommand{\pna}{}\renewcommand{\pna}{\textcolor{black!40}{--}}                   %
\providecommand{\rottag}{}\renewcommand{\rottag}[2]{\multirow{-#1}{*}{\rotatebox[origin=c]{90}{\scriptsize\textit{#2}}}}

\newcommand{\TOneSetup}{%
  \ifTOneProps\def\TP##1{##1}\else\def\TP##1{}\fi
  \ifTOneAvg\def\TA##1{##1}\else\def\TA##1{}\fi
  \edef\TOneFirst{\ifTOneProps 5\else 3\fi}%
  \edef\TOneLast{\the\numexpr\TOneFirst+6\ifTOneAvg+1\fi\relax}%
  \edef\TOneBand{\the\numexpr\TOneLast-1\relax}%
  \edef\TOneSpec{@{}c l\ifTOneProps\space c c\fi\space ccc c cc c\ifTOneAvg\space c\fi @{}}%
  \edef\TOneCmid{%
    \noexpand\cmidrule(lr){\TOneFirst-\the\numexpr\TOneFirst+2\relax}%
    \noexpand\cmidrule(lr){\the\numexpr\TOneFirst+3\relax-\the\numexpr\TOneFirst+3\relax}%
    \noexpand\cmidrule(lr){\the\numexpr\TOneFirst+4\relax-\the\numexpr\TOneFirst+5\relax}%
    \noexpand\cmidrule(lr){\the\numexpr\TOneFirst+6\relax-\the\numexpr\TOneFirst+6\relax}}%
  \edef\TOneCmidOurs{\noexpand\cmidrule(l){2-\TOneLast}}%
  \ifTOneProps\def\TOneDeltaLabel##1{\multicolumn{3}{r}{##1}}\else\def\TOneDeltaLabel##1{##1}\fi
  \def\famrow##1{& \multicolumn{\TOneBand}{l}{\cellcolor{TOneFam}\textit{##1}} \\}%
}
\newcommand{\beginTOne}{\edef\TOneTmp{\noexpand\begin{tabular}{\TOneSpec}}\TOneTmp}

\newif\ifTTwoWS   \TTwoWStrue    %
\newif\ifTTwoUSD  \TTwoUSDtrue   %

\providecolor{OursBg}{HTML}{FFF8E7}
\providecolor{forestgreen}{HTML}{228B22}
\providecolor{darkred}{HTML}{B3412C}
\providecolor{phgray}{HTML}{A0A0A0}
\providecolor{TOneQA}{HTML}{DDE1F1}
\providecolor{TOneCode}{HTML}{E0EBD7}
\providecolor{TOneWeb}{HTML}{FDF2D1}
\providecolor{TOneFam}{HTML}{F4F4F4}
\definecolor{TTwoTrain}{HTML}{F7E4E1}
\providecolor{OldMark}{HTML}{B06A00}

\providecommand{\ddown}{}\renewcommand{\ddown}[1]{\textcolor{darkred}{$-$#1}}
\providecommand{\oc}{}\renewcommand{\oc}{\cellcolor{OursBg}}

\newcommand{\TTwoSetup}{%
  \ifTTwoWS\def\TW##1{##1}\else\def\TW##1{}\fi
  \ifTTwoUSD\def\TU##1{##1}\else\def\TU##1{}\fi
  \edef\TTwoNhq{\ifTTwoUSD 5\else 4\fi}%
  \edef\TTwoNmb{\ifTTwoUSD 4\else 3\fi}%
  \edef\TTwoNws{\ifTTwoUSD 3\else 2\fi}%
  \edef\TTwoHQb{\the\numexpr 5+\TTwoNhq\relax}%
  \edef\TTwoMBa{\the\numexpr\TTwoHQb+1\relax}%
  \edef\TTwoMBb{\the\numexpr\TTwoHQb+\TTwoNmb\relax}%
  \edef\TTwoWSa{\the\numexpr\TTwoMBb+1\relax}%
  \edef\TTwoWSb{\the\numexpr\TTwoMBb+\TTwoNws\relax}%
  \edef\TTwoLast{\ifTTwoWS\TTwoWSb\else\TTwoMBb\fi}%
  \edef\TTwoSpec{@{}l l r r c\space c c c c\ifTTwoUSD\space c\fi\space c c c\ifTTwoUSD\space c\fi\ifTTwoWS\space c c\ifTTwoUSD\space c\fi\fi @{}}%
  \edef\TTwoCmid{\noexpand\cmidrule(lr){3-4}\noexpand\cmidrule(lr){6-\TTwoHQb}%
    \noexpand\cmidrule(lr){\TTwoMBa-\TTwoMBb}\ifTTwoWS\noexpand\cmidrule(lr){\TTwoWSa-\TTwoWSb}\fi}%
  \edef\TTwoCmidTok{\noexpand\cmidrule(lr){8-9}}%
  \edef\TTwoCmidOurs{\noexpand\cmidrule(lr){1-\TTwoLast}}%
  \def\famrowII##1{\multicolumn{\TTwoLast}{l}{\cellcolor{TOneFam}\textit{##1}} \\}%
}
\newcommand{\beginTTwo}{\edef\TTwoTmp{\noexpand\begin{tabular}{\TTwoSpec}}\TTwoTmp}

\providecommand{\yes}{}\renewcommand{\yes}{\textcolor{forestgreen}{\ding{51}}}        %
\providecommand{\no}{}\renewcommand{\no}{\textcolor{darkred}{\ding{55}}}              %

\definecolor{OODMark}{HTML}{D9531E}
\newcommand{\ood}{\textsuperscript{\textcolor{OODMark}{\fontsize{3.8}{3.8}\selectfont\textsf{OOD}}}}

\title{RoleFlow: Self-Evolving Multi-Role Agentic Collaboration \\ via Structural Adaptation}

\author{
  Yi Ge\textsuperscript{1} \quad
  Pan Wang\textsuperscript{2} \quad
  Xuehao Cui\textsuperscript{1} \quad
  Zian Zeng\textsuperscript{1} \\
  \textbf{Xu Hu\textsuperscript{3}} \quad
  \textbf{Heng Huang\textsuperscript{1,\dag}} \quad
  \textbf{Furong Huang\textsuperscript{1,\dag}} \\[0.6ex]
  \textsuperscript{1}University of Maryland, College Park \quad
  \textsuperscript{2}University of Pittsburgh \\
  \textsuperscript{3}The University of Texas at Dallas
}

\begin{document}

\maketitle
{\renewcommand{\thefootnote}{}%
 \footnotetext{\raggedright\textsuperscript{\dag}Corresponding authors.\\
 \texttt{\{yige,\,furongh,\,heng\}@umd.edu}}}

\begin{abstract}

Large language model (LLM) based multi-agent systems often rely on predefined collaboration structures, limiting their adaptability to evolving task requirements. Although recent approaches enable dynamic orchestration, trajectory-level rewards they apply assign one outcome to each decision, ofen struggling to identify the contributions of individual structural changes. We propose \textbf{RoleFlow}, a self-evolving multi-role agentic system that refines collaboration structures through feedback-driven atomic edits and on-demand role synthesis. To make structural adaptation learnable, we introduce \textbf{Single-Edit Counterfactual Policy Optimization (SECPO)}, which evaluates each edit against its no-edit counterpart at the same chain state and assigns targeted credit to edit-specific tokens.
Across seven benchmarks spanning four domains, RoleFlow achieves the optimal average performance, surpassing the strongest competitor by 4.7 points. It leads on all three multi-hop QA benchmarks by 3.2--7.5 F1 points and also shows supremacy on mathematical reasoning, web interaction, and code generation. Further analyses demonstrate cross-benchmark transferability, training-data efficiency, and cross-executor generalization.

\end{abstract}

\section{Introduction}

\begin{figure}[!t]
  \centering
  \includegraphics[width=\columnwidth]{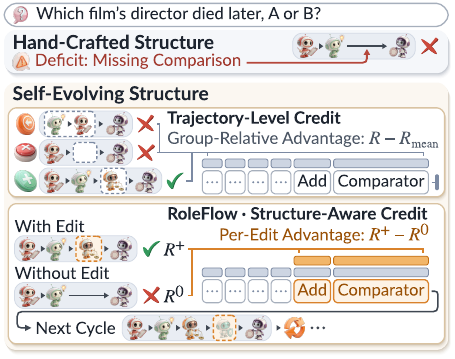}
  \caption{Motivation of RoleFlow. Hand-crafted structures lack adaptability \textbf{(top)}, while trajectory-level rewards obscure the contributions of individual structural edits \textbf{(middle)}. Our RoleFlow enables effective structural evolution through same-state counterfactual evaluation and edit-level credit assignment \textbf{(bottom)}.}
  \label{fig:secpo_anatomy}
  \vspace{-0.5em}
\end{figure}

\label{sec:introduction}

Large language model (LLM) based multi-agent systems have shown capabilities in solving complex tasks through specialized agent collaboration~\cite{durante2025interactive,ghareeb2026multi}. However, existing approaches typically rely on predefined or pre-optimized collaboration structures that remain fixed during execution. Such structures struggle to accommodate evolving task requirements, where effective collaboration may demand adding, removing, or reorganizing specialized roles. This raises a fundamental question: \emph{How can multi-role agentic systems autonomously evolve their collaboration structures during task execution?}

Recent studies have explored reinforcement learning for dynamic agent orchestration \citep{li2025agentflow,zhang2026flowsteer,nie2025w4s}. However, learning structural adaptation introduces a critical challenge: \emph{credit assignment for individual structural decisions}. Conventional methods like Group Relative Policy Optimization (GRPO) \citep{shao2024deepseekmath} rely on trajectory-level rewards, assigning the same advantage to decisions within a rollout regardless of their individual contributions. As shown in Fig.\,\ref{fig:secpo_anatomy}, when collaboration structures undergo successive modifications, this coarse-grained feedback cannot reliably identify which edits improve performance and which undermine it.

\textbf{Our key insight} is that reliable structural evolution requires making each structural modification individually assessable and learnable. Rather than treating an evolving collaboration structure as a sequence of decisions optimized only through its final outcome, we decompose structural adaptation into atomic edits and evaluate each edit against an unchanged counterpart under the same chain state. This isolates the immediate contribution of a structural decision and provides a localized learning signal, allowing the system to learn not only \emph{how to modify} collaboration structures but also \emph{which modifications are worth making}.

Motivated by this insight, we propose \textbf{RoleFlow}, a self-evolving multi-role collaboration framework that continuously adapts agent structures through execution feedback. A trainable \textbf{Global Router (GR)} constructs task-specific role chains and refines them through atomic structural edits, while synthesizing specialized roles on demand. To optimize these decisions, we introduce \textbf{Single-Edit Counterfactual Policy Optimization (SECPO)}, which compares executions with and without each candidate edit under the same chain state and incorporates the resulting advantage into edit-specific token optimization. RoleFlow further maintains a return-gated role library to retain useful synthesized specialists during training. Experiments across seven benchmarks spanning four domains demonstrate the effectiveness of RoleFlow, achieving the highest average performance among evaluated multi-agent systems.

\noindent\textbf{Our contributions are threefold:}
\begin{itemize}
    \item We introduce \textbf{RoleFlow}, a self-evolving multi-role agentic system that enables feedback-driven structural adaptation through atomic edits and on-demand specialist synthesis.
    
    \item We propose \textbf{SECPO}, a structure-aware policy optimization method that estimates the counterfactual contribution of individual structural edits and assigns targeted credit to edit-specific tokens.

\item Experiments across seven benchmarks spanning four domains show that RoleFlow achieves the highest average performance. Further analyses validate structural adaptation, edit-level counterfactual learning, cross-benchmark transferability, training-data efficiency, and cross-executor generalization.

\end{itemize}

\section{Related Work}
\label{sec:related_work}

\paragraph{Handcrafted Multi-Agent Structure and Content-Level Adaptation.}
Existing multi-agent systems often rely on fixed collaboration structures, such as the predefined pipeline in ChatDev~\citep{qian2024chatdev}, sequential agents in Chain-of-Agents~\citep{zhang2024chain}, and graph-based retrieval in LinearRAG~\citep{linearrag2025}. Other approaches adapt the information processed within a fixed structure through retrieval or tool use~\citep{lewis2021retrievalaugmentedgenerationknowledgeintensivenlp,schick2023toolformerlanguagemodelsteach,song2025r1searcher}. While effective at content-level adaptation, these methods do not learn to modify agent roles or their interactions in response to execution feedback.

\paragraph{Self-Evolving Agents and Workflow Search.}
Existing self-evolving agents primarily optimize model behavior within fixed roles, as in Multi-Agent Evolve~\citep{ulab2025mae}. Alternatively, workflow optimization methods search for effective collaboration structures through Monte Carlo tree search (AFlow)~\citep{zhang2025aflow}, learned graph connections (GPTSwarm)~\citep{zhuge2024gptswarm}, offline workflow generation (W4S)~\citep{nie2025w4s}, or task-specific architecture generation (MaAS and MAS-GPT)~\citep{zhang2025maas,ye2025masgpt}. However, these methods determine the collaboration structure before execution, limiting their ability to address emerging task requirements through online structural refinement.

\paragraph{Online Structural Adaptation and Credit Assignment.}
Recent approaches incorporate execution feedback into agent orchestration, including AgentFlow's learned planner~\citep{li2025agentflow}, AgentConductor's dynamic topology~\citep{wang2025agentconductor}, and FlowSteer's atomic workflow edits~\citep{zhang2026flowsteer}. However, their trajectory-level rewards provide coarse-grained credit across structural decisions~\citep{shao2024deepseekmath}. Counterfactual credit assignment has been explored in COMA~\citep{foerster2018counterfactual}, while COLLATOR~\citep{chen2026collator}, SERO~\citep{ge2026sero}, and EvoSteer~\citep{zhang2026evosteer} evaluate individual decisions using different references. Unlike these approaches, \textbf{RoleFlow} evaluates each structural edit against its no-edit counterpart at the same chain state. Its \textbf{SECPO} assigns the resulting counterfactual advantage specifically to edit-related tokens, enabling fine-grained credit assignment for evolving collaboration structures.

\section{Approach}
\label{sec:methods}

\begin{figure*}[!t]
  \centering
  \includegraphics[width=\linewidth]{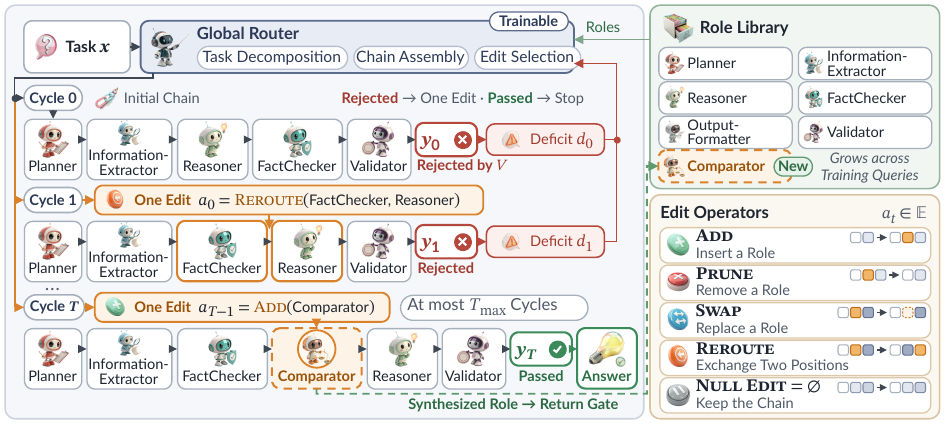}
  \caption{\textbf{Overview of RoleFlow.} The Global Router assembles an initial sequence of roles from the role library $\RoleLib$ and applies one structural edit $a_t \in \mathbb{E}$ upon validation failure, for at most $T_{\max}$ adaptation cycles. The role library grows through the selective retention of synthesized roles during training.}
  \label{fig:roleflow_overview}
\end{figure*}

As illustrated in Figure\,\ref{fig:roleflow_overview}, RoleFlow is a self-evolving multi-role agentic system that adapts collaboration structures during task execution. Its Global Router (GR) selects and revises a chain of specialized roles using validation feedback. To make these structural decisions learnable, we introduce Single-Edit Counterfactual Policy Optimization (SECPO), which assigns each candidate edit a same-state counterfactual advantage (Figure~\ref{fig:secpo_advantage}). For the subsequent content, we first define the structural adaptation problem, then describe RoleFlow's execution loop, SECPO, and the mechanism for synthesizing and retaining roles.

\subsection{Problem Formulation}
\label{sec:problem-setup}

For a task $x\in\mathcal{X}$, RoleFlow uses a trainable Global Router (GR) with parameters $\theta$ and a role library $\RoleLib=\{\rho^{(j)}\}_{j=1}^{|\RoleLib|}$. Each role $\rho$ specifies a name, a system prompt, and capability tags. A RoleFlow chain $\RFC_t$ is an ordered sequence of roles selected from the available library. A frozen executor LLM runs these roles sequentially, with each role receiving the task and the outputs of preceding roles. By default, the chain begins with a Planner and ends with a Validator.

At cycle $t$, executing $\RFC_t$ produces an output $y_t$, execution trace $z_t$, and cost $c_t$. A separate task validator $V$ assesses the output and trace, returning a pass/reject indicator $h_t\in\{0,1\}$, score $s_t$, and deficit description $d_t$. Scratch memory $\sigma_t$ stores the execution traces and validation history. Together, the chain, memory, and available roles constitute the current chain state.

When validation fails, the GR selects a structural edit $a_t\in\mathbb{E}$, where $\mathbb{E}$ contains $\textsc{Add}(\rho)$ (insert $\rho$ before the final role), $\textsc{Prune}(\rho)$ (remove $\rho$), $\textsc{Swap}(\rho,\rho')$ (replace $\rho$ by $\rho'$ in place), $\textsc{Reroute}(p,q)$ (exchange the roles at positions $p$ and $q$), and the null edit $\varnothing$ (leave the chain unchanged, $\RFC_t\oplus\varnothing=\RFC_t$). The execution and adaptation cycle is
\begin{equation}
\begin{aligned}
(y_t, z_t, c_t) &= \RFC_t(\sigma_t, x),\\
(h_t, s_t, d_t) &= V(y_t, z_t),\\
a_t &\sim \pi_{\GR,\theta}(\,\cdot \mid \RFC_t, \sigma_t, d_t),\;\; a_t \in \mathbb{E},\\
\RFC_{t+1} &= \RFC_t \oplus a_t.
\end{aligned}
\label{eq:cycle-dynamics}
\end{equation}

The initial chain is executed at cycle $0$, followed by at most $T_{\max}$ adaptation cycles, for a maximum of $1+T_{\max}$ chain executions at inference. An edit is proposed only after rejection, and each cycle applies at most one edit. This atomicity is important for measuring the effect of an individual structural decision in SECPO (\S\ref{sec:secpo-objective}).

\paragraph{Learning objective.}
The executor remains frozen; only the GR is trained. We maximize $\mathbb{E}_{x}[R(\tau_\theta(x))]$ over $\theta$, where $\tau_\theta(x)$ is a rollout of Eq.~\ref{eq:cycle-dynamics}. Its terminal cycle $T$ is the first passing cycle, or $T_{\max}$ if none passes. The cycle reward is $R(y_t)=h_t+\lambda_s s_t$, with score weight $\lambda_s$, and the rollout reward is $R(\tau)=R(y_T)$. Because this terminal reward evaluates the entire evolving chain, it does not reveal  each edit's contribution. We use $K$ rollouts per task, indexed by $i\in\{0,\ldots,K{-}1\}$; $t$ indexes cycles, and $\ell$ indexes GR plan tokens.

\subsection{RoleFlow: Feedback-Driven Structural Adaptation}
\label{sec:roleflow-runtime}

RoleFlow evolves a chain within a task rather than fixing its structure before execution. The GR first writes a plan that decomposes the task and names the roles of $\RFC_0$. After each rejected output, it uses the deficit $d_t$ and accumulated memory $\sigma_t$ to propose one structural edit. The edited chain is then executed in the next cycle, and the process stops upon validation success or after $T_{\max}$ adaptation cycles (Figure~\ref{fig:roleflow_overview}). This one-edit-per-cycle design ensures that consecutive chains differ by at most one edit; applying several edits at once would confound their individual effects.

The available roles are not restricted to a fixed predefined set. If the GR names a role absent from $\RoleLib$, a LibraryExpander synthesizes its specification from the role name, task, and current deficit. The new role is stored in rollout $i$'s temporary overlay $\hat{\RoleLib}^{(i)}$, so that rollout can use roles from $\RoleLib\cup\hat{\RoleLib}^{(i)}$. Moreover, if the GR proposes a null edit despite a reported deficit, RoleFlow synthesizes a role targeting that deficit and proposes its addition (\emph{deficit-driven synthesis}). We describe how synthesized roles are retained across training steps in \S\ref{sec:library}.

At inference, each task starts with the domain's seed library rather than the library grown during training, and the required specialists are synthesized for that task. Each chain is executed once per cycle; the additional no-edit executions described next are used only during SECPO training.

\subsection{SECPO: Edit-Level Counterfactual Credit Assignment}
\label{sec:secpo-objective}

\begin{figure}[t]
  \centering
  \includegraphics[width=\columnwidth]{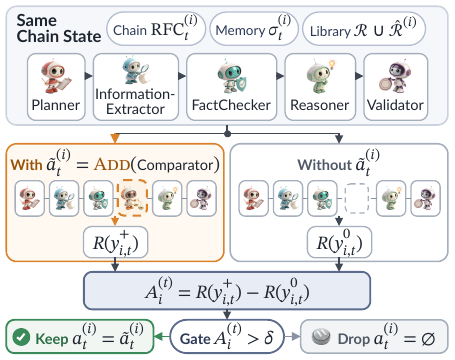}
  \caption{The single-edit advantage of SECPO. A certain chain state is executed with and without the candidate edit, and the reward difference $A^{(t)}_i$ (Eq.~\ref{eq:secpo-adv}) credits that edit and gates its commitment, enabling edit-level credit assignment under evolving structural dynamics.}
  \label{fig:secpo_advantage}
\end{figure}

Structural adaptation presents a credit-assignment problem: the terminal reward reflects the outcome of several decisions, even though an individual edit may improve or worsen the chain. GRPO's group-relative advantage also mixes rollouts with different structures, while a learned value critic would need to track changing chain states. Counterfactual methods such as COMA~\citep{foerster2018counterfactual} require evaluating alternative actions at each step. SECPO instead measures a candidate edit against the unchanged chain at the \emph{same} state, while retaining GRPO's critic-free group of $K$ rollouts~\citep{shao2024deepseekmath} and PPO's clipped surrogate with KL regularization~\citep{schulman2017proximal}.

\paragraph{Same-state counterfactual advantage.}
For each task, as shown in Figure\,\ref{fig:secpo_advantage}, sample $K$ rollouts $\{\tau^{(i)}\}_{i=0}^{K-1}\sim\pi_{\GR,\theta_{\mathrm{old}}}(\cdot\mid x)$, where $\theta_{\mathrm{old}}$ denotes the policy before an update. Each rollout maintains its own role overlay $\hat{\RoleLib}^{(i)}$ until the training step ends. At cycle $t$, the candidate edit $\tilde a_t^{(i)}$ is either the GR's proposed edit or the $\textsc{Add}$ operation produced by deficit-driven synthesis (\S\ref{sec:roleflow-runtime}). Keeping $\RFC_t^{(i)}$ and $\sigma_t^{(i)}$ fixed, SECPO executes the chain with the candidate edit, $(\RFC_t^{(i)}\oplus\tilde a_t^{(i)})(\sigma_t^{(i)},x)$, to obtain $y_{i,t}^{+}$, and without it, $\RFC_t^{(i)}(\sigma_t^{(i)},x)$, to obtain $y_{i,t}^{0}$. The resulting single-edit advantage is
\begin{equation}
A^{(t)}_i \,=\, R(y^{+}_{i,t}) - R(y^{0}_{i,t}).
\label{eq:secpo-adv}
\end{equation}

This reward difference provides a structure-aware estimate of the candidate edit's immediate effect at the current chain state. SECPO also uses it to decide whether the edit is committed during training: only edits whose advantage exceeds a positive threshold $\delta$ are retained,

\begin{equation}
a^{(i)}_t = \begin{cases} \tilde a^{(i)}_t, & A^{(t)}_i > \delta \\ \varnothing, & \text{otherwise}, \end{cases}
\label{eq:greedy-commit}
\end{equation}

Otherwise, the chain remains unchanged. Importantly, the counterfactual pair contributes to the policy update even when the candidate edit is not committed.

\begin{figure}[t]
  \centering
  \includegraphics[width=\columnwidth]{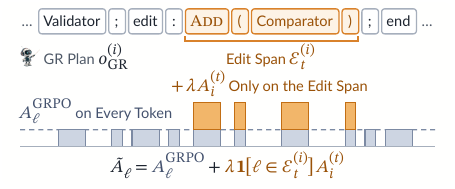}
  \caption{Token-level advantage of SECPO. Every plan token receives the group-relative advantage $A^{\mathrm{GRPO}}_\ell$, and the single-edit advantage $A^{(t)}_i$ is added with weight $\lambda$ on the edit span $\mathcal{E}^{(i)}_t$.}
  \label{fig:token_level}
\end{figure}

\paragraph{Edit-specific token advantage.}
\label{sec:token-mask}
The group-relative GRPO advantage $A_\ell^{\mathrm{GRPO}}$ is the terminal rollout return $R(\tau^{(i)})$, standardized across the task's $K$ rollouts and broadcast to all GR plan tokens. This signal does not distinguish the tokens corresponding to different structural edits. Let $\mathcal{E}_t^{(i)}$ denote the plan-token span spelling candidate edit $\tilde a_t^{(i)}$. SECPO adds the single-edit advantage only on that span (Figure~\ref{fig:token_level}):

\begin{equation}
\tilde A_\ell \,=\, A^{\mathrm{GRPO}}_\ell + \lambda\,\mathbf{1}\!\left[\ell \in \mathcal{E}^{(i)}_t\right] A^{(t)}_i ,
\label{eq:tokenmask}
\end{equation}
where $\lambda$ controls the edit-specific signal. The spans from different adaptation cycles do not overlap, and tokens outside all edit spans retain their GRPO advantages. We set $\lambda=1$; the GRPO ablation in Table~\ref{tab:ablation} uses $\lambda=0$ and omits no-edit re-execution.

\paragraph{Policy optimization.}
Using the token-level advantage $\tilde A_\ell$, we optimize the GR with PPO's clipped objective, averaged over the $K$ rollouts and their GR plan tokens $o_{\GR}^{(i)}$:

\begin{equation}
\begin{multlined}
\mathcal{L}^{\text{SECPO}}(\theta) = -\,\frac{1}{K}\sum_{i=0}^{K-1}\frac{1}{|o^{(i)}_{\GR}|}\sum_{\ell=1}^{|o^{(i)}_{\GR}|}\\
\min\!\big(\rho_\ell\,\tilde A_\ell,\ \mathrm{clip}(\rho_\ell,1{-}\epsilon,1{+}\epsilon)\,\tilde A_\ell\big)\\ \,+\,\beta\,D_{\mathrm{KL}}(\pi_\theta\,\|\,\pi_{\mathrm{ref}}),
\end{multlined}
\label{eq:secpo-loss}
\end{equation}
where $\rho_\ell(\theta)=\pi_\theta(o_\ell\mid o_{<\ell})/\pi_{\theta_{\mathrm{old}}}(o_\ell\mid o_{<\ell})$ is the per-token probability ratio, $\pi_{\mathrm{ref}}$ is the reference policy, $\epsilon$ is the clipping parameter, and $\beta$ is the KL coefficient (\S\ref{sec:exp_setup}), and $\pi_\theta$ abbreviates $\pi_{\GR,\theta}$. Algorithm~\ref{alg:roleflow-secpo-detailed} in Appendix~\ref{app:secpo-full-algorithm} summarizes the Global Router learning with SECPO.

\subsection{On-Demand Role Synthesis and Library Growth}
\label{sec:library}

RoleFlow uses temporary role overlays for task-specific specialists and a shared library for roles retained across training steps. After each training step, a \emph{return gate} determines which synthesized role specifications enter the shared library. A synthesized role $\hat\rho$ from rollout $i$ is admitted only if its rollout return exceeds that of reference rollout $\tau_{\mathrm{ref}}^{(i)}$, defined as the first rollout in its block of $K$ within the training batch (Algorithm~\ref{alg:roleflow-secpo-detailed}). Thus, after training step $s$,

\begin{equation}
\small
\RoleLib_{s+1} = \RoleLib_s \cup \{\hat\rho \mid \exists\, i:\hat\rho\in\hat{\RoleLib}^{(i)},\, R(\tau^{(i)}) > R(\tau^{(i)}_{\mathrm{ref}})\}.
\label{eq:library-commit}
\end{equation}

The shared library is then available to subsequent training steps. This gate applies to role specifications rather than gradients: even if a synthesized role is not retained in the library, its execution rewards still contribute to GR training. At inference, RoleFlow uses the domain's seed roles and synthesizes missing specialists on demand, without loading the training-grown library.

\section{Experiments}
\label{sec:experiments}

We investigate whether feedback-driven structural evolution improves multi-agent collaboration, which components make this evolution effective, and whether the learned adaptation transfers across tasks and executors. 
In particular, we assess overall performance (\S\ref{sec:main_results}), isolate key components and credit signals (\S\ref{sec:ablations}), and examine transfer, efficiency, and executor generalization (\S\ref{sec:role_library_main}; \S\ref{sec:efficiency}).

\begin{table*}[t]
\centering
\caption{Performance comparison with direct prompting, hand-crafted pipelines, automated workflow search, trained executors and trained orchestrators on seven benchmarks. The executor LLM is consistently set as gpt-4o-mini for all multi-agent systems and trained orchestrators. We bold the best results and underline the runner-ups among all methods for direct prompting and remaining methods, respectively; \ood{} marks out-of-distribution benchmarks.}
\label{tab:main_6domain}\label{tab:main}
\scriptsize
\setlength{\tabcolsep}{4.5pt}
\renewcommand{\arraystretch}{1.0}
\rowcolors{2}{gray!11}{white}
\setlength{\aboverulesep}{0.15ex}\setlength{\belowrulesep}{0.25ex}%
\setlength{\heavyrulewidth}{1.1pt}%
\def\TP#1{}%
\def\famrow#1{\hiderowcolors & \multicolumn{9}{c}{\textcolor{black!55}{\textit{#1}}} \\ \cmidrule{2-10}\showrowcolors}%
\def\tg{\cellcolor{white}}%
\providecolor{OODMark}{HTML}{D9531E}\providecommand{\ood}{\textsuperscript{\textcolor{OODMark}{\fontsize{3.8}{3.8}\selectfont\textsf{OOD}}}}%
\providecommand{\rottag}{}\renewcommand{\rottag}[2]{\multirow{-#1}{*}{\rotatebox[origin=c]{90}{\scriptsize\textit{#2}}}}%
\resizebox{0.9\textwidth}{!}{%
\begin{tabular}{@{}c@{\hspace{3pt}}l|ccc|c|cc|c!{\color{black!35}\vrule}c@{}}
\toprule
\hiderowcolors
 & \multirow{2}{*}{\textbf{Method}} & \multicolumn{3}{c|}{\cellcolor{TOneQA}\textbf{Multi-hop QA}} & \cellcolor{TOneReason}\textbf{Reasoning} & \multicolumn{2}{c|}{\cellcolor{TOneCode}\textbf{Coding}} & \cellcolor{TOneWeb}\textbf{Web} & \cellcolor{TOneAvg} \\
 & & HotpotQA & 2Wiki\ood & MuSiQue\ood & MATH-500 & MBPP & HumanEval\ood & WebShop & \cellcolor{TOneAvg}\textbf{Avg.} \\
\midrule
\showrowcolors
\famrow{Direct Prompting}
\tg & GPT-4o-mini \TP{& \pna & \pna} & \second{45.6} & 24.5 & \second{23.4} & 76.5 & 83.7 & 88.3 & 59.0 & \ac57.3 \\
\tg & Claude-Haiku-4.5 \TP{& \pna & \pna} & \best{47.2} & \best{26.4} & \best{28.1} & \second{92.0} & \best{90.6} & \best{96.9} & 59.5 & \ac\best{63.0} \\
\tg & Qwen2.5-7B-Instruct \TP{& \pna & \pna} & 30.5 & 18.8 & 11.6 & 77.5 & 76.8 & 82.4 & 58.9 & \ac50.9 \\
\tg & Llama-3.1-8B-Instruct \TP{& \pna & \pna} & 32.6 & 15.1 & 9.5 & 53.3 & 70.7 & 71.8 & 57.2 & \ac44.3 \\
\tg & Qwen3-8B \TP{& \pna & \pna} & 31.2 & \second{25.2} & 13.5 & 83.9 & 78.3 & 87.0 & 62.0 & \ac54.4 \\
\tg & Qwen3-32B \TP{& \pna & \pna} & 30.5 & 24.5 & 16.6 & 84.5 & 83.3 & 90.1 & \best{64.7} & \ac56.3 \\
\tg & DeepSeek-V3.2 \TP{& \pna & \pna} & 44.1 & 19.7 & 21.2 & \best{95.4} & \second{88.3} & \second{94.9} & \second{63.3} & \ac\second{61.0} \\
\cmidrule{2-10}
\famrow{Hand-Crafted Pipeline}
\tg & CoA \TP{& \pno & \pno} & 44.6 & 25.9 & 17.7 & 74.6 & 81.0 & 87.0 & 53.4 & \ac54.9 \\
\tg & LinearRAG \TP{& \pna & \pna} & 47.5 & 33.3 & 32.8 & 77.0 & 84.5 & 50.4 & 56.4 & \ac54.6 \\
\cmidrule{2-10}
\famrow{Automated Workflow Search}
\tg & GPTSwarm \TP{& \pno & \pno} & 48.5 & 26.7 & 21.1 & 79.2 & 86.9 & 91.6 & 58.7 & \ac59.0 \\
\tg & AFlow \TP{& \pno & \pno} & 50.1 & 31.3 & 31.1 & 79.2 & 88.8 & 95.4 & \second{59.3} & \ac62.2 \\
\tg\rottag{15}{Training-free} & MaAS \TP{& \pyes & \pno} & 47.9 & 31.6 & 29.1 & 77.4 & 88.6 & \best{96.2} & 55.2 & \ac60.9 \\
\midrule\midrule
\famrow{Trained Executor}
\tg & MAE \TP{& \pna & \pna} & 16.7 & 17.8 & 9.1 & 64.8 & 64.2 & 66.4 & 53.6 & \ac41.8 \\
\tg & R1-Searcher \TP{& \pna & \pna} & 50.3 & \second{39.9} & \second{34.6} & 76.0 & 78.9 & 84.0 & 55.9 & \ac59.9 \\
\cmidrule{2-10}
\famrow{Trained Orchestrator}
\tg & AgentFlow \TP{& \pyes & \pyes} & 37.8 & 32.3 & 28.2 & 70.8 & 78.6 & 84.0 & 56.7 & \ac55.5 \\
\tg & FlowSteer \TP{& \pyes & \pyes} & 43.4 & 24.0 & 21.7 & 67.6 & 82.7 & 82.4 & 50.3 & \ac53.2 \\
\tg & MAS-GPT \TP{& \pyes & \pno} & 41.4 & 24.0 & 27.3 & 79.6 & 83.9 & 91.6 & 54.9 & \ac57.5 \\
\tg & W4S \TP{& \pno & \pno} & \second{52.4} & 35.7 & 28.8 & \second{80.0} & \best{89.7} & 93.1 & 57.1 & \ac\second{62.4} \\
\cmidrule(lr){2-10}
\rowcolor{OursBg} \tg\rottag{9}{Post-training} & \textbf{\textsc{RoleFlow}} \TP{& \pyes & \pyes} & \best{55.6}\sd{2.2} & \best{43.5}\sd{1.8} & \best{42.1}\sd{1.4} & \best{81.1}\sd{0.5} & \second{89.0}\sd{0.2} & \second{95.7}\sd{0.9} & \best{62.6}\sd{1.2} & \oc\best{67.1} \\
\bottomrule
\end{tabular}}
\end{table*}

\subsection{Experimental Setup}
\label{sec:exp_setup}

\paragraph{Datasets.}
We evaluate RoleFlow across four task domains: \textbf{(1) Multi-hop QA} with HotpotQA~\citep{yang2018hotpotqa}, 2WikiMultiHopQA~\citep{ho2020constructing} and MuSiQue~\citep{trivedi2022musique}; \textbf{(2) Mathematical reasoning} with MATH-500~\citep{lightman2023letsverify}; \textbf{(3) Code generation} with MBPP~\citep{austin2021mbpp} and HumanEval~\citep{chen2021humaneval}; and \textbf{(4) Web interaction} with WebShop~\citep{yao2022webshop} under the AgentBench protocol~\citep{liu2024agentbench}. We train a domain-specific GR on the HotpotQA, MATH~\citep{hendrycks2021math}, MBPP, and WebShop training sets. We evaluate 2Wiki, MuSiQue, and HumanEval without retraining, using the GR trained on HotpotQA or MBPP, respectively. Dataset details appear in Appendix~\ref{sec:appx_data}.

\paragraph{Baselines.}
Table~\ref{tab:main} compares four categories of baselines: \textbf{(1) Hand-crafted pipelines}: Chain-of-Agents (CoA)~\citep{zhang2024chain} and LinearRAG~\citep{linearrag2025}; \textbf{(2) Automated workflow search}: GPTSwarm~\citep{zhuge2024gptswarm}, AFlow~\citep{zhang2025aflow} and MaAS~\citep{zhang2025maas}; \textbf{(3) Trained executors}: MAE~\citep{ulab2025mae} and R1-Searcher~\citep{song2025r1searcher}; and \textbf{(4) Trained orchestrators}: AgentFlow~\mbox{\citep{li2025agentflow}}, FlowSteer~\citep{zhang2026flowsteer}, W4S~\citep{nie2025w4s} and MAS-GPT~\citep{ye2025masgpt}. Every multi-agent system and trained orchestrator executes its agents with gpt-4o-mini~\citep{openai2024gpt4o}; R1-Searcher, AgentFlow and LinearRAG retrieve from Wikipedia. We also report \textbf{direct prompting} of GPT-4o-mini, Claude-Haiku-4.5~\citep{anthropic2025haiku45}, Qwen2.5-7B-Instruct~\citep{qwen2024qwen25}, Llama-3.1-8B-Instruct~\citep{dubey2024llama3}, Qwen3-8B, Qwen3-32B~\citep{yang2025qwen3} and DeepSeek-V3.2~\citep{deepseek2025v32} with each benchmark's official evaluation protocol (Appendix~\ref{sec:appx_direct_prompt}).

\paragraph{Metric.}
We report F1 for QA, accuracy for MATH-500, Pass@1 for MBPP and HumanEval, and the WebShop score. Results are means and standard deviations over three training runs.

\paragraph{Implementation.}
We train the GR with SECPO using Qwen2.5-7B-Instruct in verl~\citep{sheng2024hybridflow} on four NVIDIA H200 GPUs; roles are executed by gpt-4o-mini. Each task uses $K{=}5$ rollouts, with 32 tasks per training step, learning rate $2\times10^{-6}$, KL coefficient 0.001, clip ratio 0.2, $T_{\max}{=}3$, and commit threshold $\delta=0.005$. Role prompts and baseline reproduction details are provided in Appendices~\ref{sec:appendix_role_library} and \ref{sec:appx_baseline_protocols}.

\subsection{Main Results}
\label{sec:main_results}

As shown in Table~\ref{tab:main}, RoleFlow achieves the highest average score of 67.1 across seven benchmarks, outperforming the strongest competing multi-agent system, W4S, by 4.7 points and ranking first on five benchmarks. Notably, RoleFlow consistently improves upon its underlying gpt-4o-mini executor across all four task domains, demonstrating the effectiveness of learned multi-role collaboration beyond individual model capabilities. The gains are particularly substantial on multi-hop QA (10.0--19.0 F1), where coordinating complementary reasoning roles is essential for handling interdependent evidence and reasoning steps. Unlike predefined or pre-optimized workflows, RoleFlow combines feedback-driven structural refinement with edit-level counterfactual learning, enabling collaboration structures to evolve in response to task requirements. These results establish the overall effectiveness of self-evolving collaboration, with subsequent ablations isolating the contributions of structural adaptation and SECPO.

\subsection{Ablation Study}
\label{sec:ablations}
\label{sec:secpo_signal}

\begin{table*}[t]
\centering
\providecommand{\GR}{\mathrm{GR}}\providecommand{\RoleLib}{\mathcal{R}}%
\providecolor{OODMark}{HTML}{D9531E}\providecommand{\ood}{\textsuperscript{\textcolor{OODMark}{\fontsize{3.8}{3.8}\selectfont\textsf{OOD}}}}%
\caption{Ablation study of RoleFlow. GR: Global Router; $\RoleLib$: role library with specialist synthesis; $T_{\max}$: adaptation cycles allowed after cycle 0 (3 in RoleFlow; 1 allows a single edit). Models are trained with $K = 5$.}
\label{tab:ablation}\label{tab:secpo_ablation}
\scriptsize
\ifdefined\abBox\else\newsavebox{\abBox}\fi%
\providecommand{\abFit}[2]{\begingroup\setlength{\tabcolsep}{3pt}\sbox{\abBox}{#2}%
 \setlength{\tabcolsep}{\dimexpr\tabcolsep+(0.66\textwidth-\wd\abBox)/\numexpr2*#1\relax\relax}\sbox{\abBox}{#2}%
 \setlength{\tabcolsep}{\dimexpr\tabcolsep+(0.66\textwidth-\wd\abBox)/\numexpr2*#1\relax\relax}#2\endgroup}
{\renewcommand{\arraystretch}{1.0}%
\abFit{9}{\begin{tabular}{ccc c ccc cc}
\toprule
\multicolumn{3}{c}{\textbf{Inference}} & \textbf{Training}
 & \multicolumn{3}{c}{\cellcolor{TOneQA}\textbf{Multi-hop QA}}
 & \multicolumn{2}{c}{\cellcolor{TOneCode}\textbf{Coding}} \\
\cmidrule(lr){1-3}\cmidrule(lr){4-4}\cmidrule(lr){5-7}\cmidrule(lr){8-9}
$\GR$ & $\RoleLib$ & $T_{\max}$ & SECPO & HotpotQA & 2Wiki\ood & MuSiQue\ood & MBPP & HumanEval\ood \\
\midrule
\oc\yes & \oc\yes & \oc 3 & \oc\yes
 & \oc\textbf{55.6}\sd{2.2} & \oc\textbf{43.5}\sd{1.8} & \oc\textbf{42.1}\sd{1.4} & \oc\underline{89.0}\sd{0.2} & \oc\textbf{95.7}\sd{0.9} \\
\no & \yes & 3 & \yes
 & 45.7\sd{1.1} & 29.6\sd{2.8} & 25.0\sd{5.0} & 88.1\sd{0.3} & 93.4\sd{0.4} \\
\yes & \no & 3 & \yes
 & 45.3\sd{1.4} & 30.6\sd{1.1} & 16.3\sd{0.6} & 86.6\sd{0.8} & 91.3\sd{0.9} \\
\yes & \yes & 1 & \yes
 & 51.8\sd{2.4} & 36.4\sd{3.5} & 33.6\sd{3.0} & 88.1\sd{0.4} & \underline{93.9}\sd{1.3} \\
\yes & \yes & 3 & \no
 & \underline{53.5}\sd{3.9} & \underline{41.0}\sd{5.0} & \underline{39.5}\sd{3.2} & \textbf{89.2}\sd{0.3} & \textbf{95.7}\sd{0.9} \\
\bottomrule
\end{tabular}}}
\end{table*}

\paragraph{Online Refinement Matters.}
Table~\ref{tab:ablation} separates initial workflow selection from subsequent adaptation. Replacing the GR with its most frequent fixed chain yields 45.7 F1 on HotpotQA, essentially the direct executor's 45.6, versus 55.6 for RoleFlow. Even with the GR retained, disabling adaptation reduces performance by 3.8--8.5 F1 across the three QA benchmarks. Thus, choosing a task-specific chain is insufficient: revising it after execution feedback accounts for a substantial part of the gain.

\paragraph{Role Synthesis Enables Adaptation.}
Removing access to synthesized roles substantially reduces QA performance (Table~\ref{tab:ablation}, disable $\RoleLib$). The resulting chains contain only one or two seed roles, with edits often collapsing to the null action. Structural revision therefore needs not only a decision about \emph{when} to edit, but also the capacity to introduce a missing specialist.

\paragraph{SECPO Improves Structural Learning.}
Replacing SECPO with GRPO~\citep{shao2024deepseekmath} lowers mean F1 by 2.1, 2.5, and 2.6 points on HotpotQA, 2Wiki, and MuSiQue, respectively (Table~\ref{tab:ablation}). GRPO also varies more across runs; one router learns to edit only once per question. This is consistent with trajectory-level credit failing to distinguish useful revisions within a rollout. The advantage is not universal: SECPO slightly trails GRPO on MBPP and ties it on HumanEval, where validation triggers fewer corrective edits.

\begin{table*}[t]
\centering
\providecommand{\Tmax}{T_{\max}}%
\providecommand{\tcpend}{\textcolor{phgray}{--}}%
\providecommand{\tcband}[1]{\hiderowcolors\multicolumn{17}{c}{\textcolor{black!55}{\textit{#1}}\global\rownum=0 } \\ \cmidrule{1-17}\showrowcolors}%
\providecommand{\tcx}[1]{#1$\times$}%
\providecolor{OODMark}{HTML}{D9531E}\providecommand{\ood}{\textsuperscript{\textcolor{OODMark}{\fontsize{3.8}{3.8}\selectfont\textsf{OOD}}}}%
\providecommand{\tctr}{\ood}%
\providecommand{\tcdup}[1]{{\fontsize{5.5}{6}\selectfont\textcolor{forestgreen}{($+$#1)}}}%
\providecommand{\tcddn}[1]{{\fontsize{5.5}{6}\selectfont\textcolor{darkred}{($-$#1)}}}%
\providecommand{\tczero}{{\fontsize{5.5}{6}\selectfont\textcolor{black!55}{(0.0)}}}%
\providecommand{\tcref}{\textcolor{black!55}{ref.}}%
\providecommand{\tcrulesep}[2]{\noalign{\global\aboverulesep=#1\global\belowrulesep=#2}}%
\providecommand{\tcpad}{\phantom{0}}%
\providecommand{\tcwrap}[2]{$\vcenter{\hbox{\strut#1}\hbox{\strut#2}}$}%
\providecommand{\tcwrapc}[2]{\setbox0=\hbox{\strut#1}\setbox2=\hbox{\strut#2}\ifdim\wd0<\wd2 \setbox0=\hbox to\wd2{\hss\unhbox0\hss}\else\setbox2=\hbox to\wd0{\hss\unhbox2\hss}\fi$\vcenter{\box0\box2}$}%
\caption{Efficiency comparison of SECPO with alternative credit signals and training budgets for the Global Router on HotpotQA. $c$: the number of chain executions per edit cycle; $\lambda$: whether the per-edit advantage is applied on the edit tokens; Bound: the maximum number of chain executions per question $= K,(1 + c,T_{\max})$; Realized: the measured mean; Rel.\ : the ratio to GRPO.}
\label{tab:secpo_compute}
\scriptsize
\setlength{\tabcolsep}{2.6pt}%
\renewcommand{\arraystretch}{1.12}
\rowcolors{1}{white}{gray!11}%
\setlength{\aboverulesep}{0.15ex}\setlength{\belowrulesep}{0.25ex}%
\setlength{\heavyrulewidth}{1.1pt}%
\begin{tabular}{l|ccc!{\color{black!35}\vrule}cc|ccc!{\color{black!35}\vrule}cc|cc!{\color{black!35}\vrule}cc!{\color{black!20}\vrule}cc}
\toprule
\hiderowcolors
\multirow{3}{*}{\textbf{Method}}
 & \multicolumn{3}{c!{\color{black!35}\vrule}}{\multirow{2}{*}{\textbf{Configuration}}}
 & \multicolumn{2}{c|}{\multirow{2}{*}{\textbf{Credit}}}
 & \multicolumn{3}{c!{\color{black!35}\vrule}}{\textbf{Chain Executions}}
 & \multicolumn{2}{c|}{\textbf{Training Run}}
 & \multicolumn{6}{c}{\cellcolor{TOneQA}\textbf{Test F1 and $\boldsymbol{\Delta}$F1 vs.\ GRPO}} \\
 & & & & &
 & \multicolumn{3}{c!{\color{black!35}\vrule}}{per Question}
 & Tokens & GPU-h
 & \multicolumn{2}{c!{\color{black!35}\vrule}}{HotpotQA} & \multicolumn{2}{c!{\color{black!20}\vrule}}{2Wiki\tctr} & \multicolumn{2}{c}{MuSiQue\tctr} \\
\tcrulesep{0pt}{0pt}%
\cmidrule(lr){7-9}\cmidrule(lr){12-13}\cmidrule(lr){14-15}\cmidrule(lr){16-17}
\tcrulesep{0.15ex}{0.25ex}%
 & $K$ & Updates & $c$ & $\lambda$ & Reference
 & Bound & Realized & Rel.
 & (M) &
 & F1 & $\Delta$ & F1 & $\Delta$ & F1 & $\Delta$ \\
\midrule
\showrowcolors
\tcband{Group-Level Credit (GRPO)}
GRPO                                     & 5 & 5  & 1 & \no & Group Mean & 20 & \tcpad9.1 & \tcx{1.00} & 4.26    & 2.6     & 53.5    & \tcref  & 41.0    & \tcref  & 39.5    & \tcref  \\
GRPO, Optimizer Updates\,$\uparrow$ & 5 & 10 & 1 & \no & Group Mean & 20 & 10.0 & \tcx{1.09} & 7.33 & 3.4 & 53.4 & \tcddn{0.1} & 40.2 & \tcddn{0.8} & 39.4 & \tcddn{0.1} \\   %
GRPO, Rollouts\,$\uparrow$ ($K = 9$) & 9 & 5 & 1 & \no & Group Mean & 36 & 16.2 & \tcx{1.78} & 7.55 & 3.5 & 54.5 & \tcdup{1.0} & 41.6 & \tcdup{0.6} & 39.3 & \tcddn{0.2} \\   %
\cmidrule{1-17}
\tcband{Per-Edit Credit without Re-Execution}
$+$\,COMA-Operator & 5 & 5 & 1 & \no & Operator Mean & 20 & \tcpad9.3 & \tcx{1.03} & 3.13 & 1.8 & 55.6 & \tcdup{2.1} & 41.4 & \tcdup{0.4} & 36.6 & \tcddn{2.9} \\   %
$+$\,Edit-Token Attrib. & 5 & 5 & 1 & \yes & Group Mean & 20 & 10.4 & \tcx{1.15} & 3.54 & 1.9 & 54.3 & \tcdup{0.8} & 39.4 & \tcddn{1.6} & 39.5 & \tczero \\   %
\cmidrule{1-17}
\tcband{Per-Edit Credit with Same-State Re-Execution (SECPO)}
\rowcolor{OursBg}\textbf{\textsc{RoleFlow}}       & 5 & 5  & 2 & \yes & No-Edit Rerun & 35 & 13.5 & \tcx{1.48} & 5.07    & 3.7     & 55.6    & \tcdup{2.1} & 43.5  & \tcdup{2.5} & 42.1  & \tcdup{2.6} \\
\cmidrule{1-17}
\tcband{Edits per Cycle (SECPO)}
Multi-Edit ($\le 2$) & 5 & 5 & 2 & \yes & No-Edit Rerun & 35 & 11.2 & \tcx{1.24} & 3.80 & 2.1 & 48.4 & \tcddn{5.1} & 37.3 & \tcddn{3.7} & 38.0 & \tcddn{1.5} \\   %
Multi-Edit ($\le 4$) & 5 & 5 & 2 & \yes & No-Edit Rerun & 35 & 10.8 & \tcx{1.19} & 3.42 & 2.1 & 50.2 & \tcddn{3.3} & 36.3 & \tcddn{4.7} & 35.9 & \tcddn{3.6} \\   %
\bottomrule
\end{tabular}
\end{table*}

\paragraph{Same-State Credit Matters.}
Table~\ref{tab:secpo_compute} distinguishes a better learning signal from a larger budget. Increasing GRPO updates or rollouts improves QA F1 by at most 1.0 point, while consuming more training tokens than SECPO. COMA-Operator matches SECPO on HotpotQA (55.6) but trails on 2Wiki and MuSiQue (41.4/36.6 versus 43.5/42.1); edit-token attribution without re-execution is likewise weaker on both OOD datasets. SECPO instead evaluates each edit against the unchanged chain at the same state (Eq.~\ref{eq:secpo-adv}), isolating a more relevant reference for edit credit and commitment (Eq.~\ref{eq:greedy-commit}). This comes at a cost: realized chain executions rise from 9.1 to 13.5 per training question relative to standard GRPO.

\paragraph{Atomic Edits Preserve Credit Granularity.}
Training with up to two or four edits per cycle lowers QA performance relative to single-edit SECPO (Table~\ref{tab:secpo_compute}). A shared counterfactual for multiple simultaneous edits measures their joint effect, obscuring individual contributions. Since these variants are evaluated with one edit per cycle, train--test mismatch also contributes to the comparison; the evidence supports the single-edit configuration without isolating that mechanism.

\subsection{Structural Adaptation and Transfer}
\label{sec:role_library_main}

\paragraph{Structure and Specialization Are Complementary.}
Preserving synthesized roles' names and positions while replacing their descriptions with generic ones retains roughly half to three-quarters of their benefit (Table~\ref{tab:library}). Much of the gain therefore comes from inserting a useful step into the chain, while tailored role instructions contribute further. Role synthesis improves both collaboration structure and specialist behavior.

\paragraph{Adaptation Transfers Across Tasks.}
The HotpotQA-trained GR transfers to 2Wiki and MuSiQue without retraining (Table~\ref{tab:main}). Its initial chains vary little, but final chains diverge almost three times as much; on MuSiQue, about a quarter of questions finish with a chain absent from the initial proposals. Almost all additional OOD specialists are synthesized after rejection (Table~\ref{tab:library}). The transferable behavior is therefore not merely an initial template, but the ability to revise collaboration in response to task feedback.

\begin{figure}[t]
  \centering
  \includegraphics[width=0.8\columnwidth]{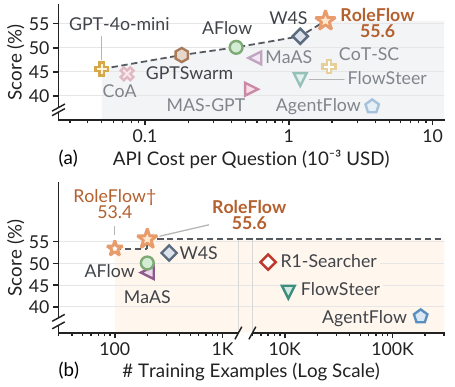}
  \caption{API Cost and data efficiency of RoleFlow on HotpotQA. Panel (a) plots F1 against the API cost per question and panel (b) against the number of labeled training questions, where RoleFlow$^{\dagger}$ denotes the GR trained on 100 labeled questions.}
  \label{fig:frontier}
\end{figure}

\subsection{Efficiency and Generalization}
\label{sec:efficiency}
\label{sec:backbone}

\paragraph{Refinement Versus Resampling.}
In Figure~\ref{fig:frontier}a, RoleFlow achieves 55.6 F1 on HotpotQA at its measured API cost, while chain-of-thought self-consistency~\citep{wang2023selfconsistency} at comparable cost remains near the executor baseline. Under this setting, investing inference in revising collaboration yields a more favorable quality--cost trade-off than sampling additional answers; it does not imply universally lower compute.

\paragraph{Training-Data Efficiency.}
As shown in Figure\,\ref{fig:frontier}b, RoleFlow achieves stronger performance with fewer labeled HotpotQA questions than competing methods. Notably, RoleFlow$^{\dagger}$, trained with half the questions, already matches the GRPO-trained router (Table~\ref{tab:ablation}). This suggests that SECPO makes more effective use of training data by extracting edit-level counterfactual signals from each task, rather than relying solely on trajectory-level rewards. Such finer-grained supervision improves data efficiency, albeit with additional training-time executions (Table~\ref{tab:secpo_compute}).

\begin{figure}[t]
  \centering
  \includegraphics[width=0.9\columnwidth]{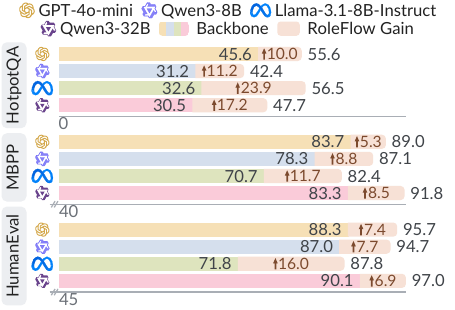}
  \caption{Performance of RoleFlow with different executor backbones on the HotpotQA, MBPP and HumanEval benchmarks.}
  \label{fig:backbone}
\end{figure}

\paragraph{Cross-Executor Generalization.}
Figure\,\ref{fig:backbone} shows that the GR trained with gpt-4o-mini improves all evaluated executors without retraining. Llama-3.1-8B-Instruct gains 11.7 and 16.0 Pass@1 on MBPP and HumanEval, compared with 5.3 and 7.4 for gpt-4o-mini. These results suggest that feedback-driven structural adaptation can compensate for weaker executors, demonstrating transferable collaboration strategies beyond executor-specific workflows.

\section{Conclusion}
\label{sec:conclusion}
We presented \textbf{RoleFlow}, a self-evolving multi-role agentic system that adapts collaboration structures through feedback-driven edits and on-demand role synthesis. We further introduced \textbf{SECPO} to enable structure-aware learning through edit-level counterfactual credit assignment. Experiments across seven benchmarks demonstrate superior average performance and strong transferability. Overall, these findings highlight the value of structure-aware credit assignment in enabling effective and adaptive multi-agent collaboration.

\section*{Limitations}
\label{sec:limitations}

RoleFlow requires a task validator that returns a pass or reject verdict and a deficit for every cycle, so a new domain needs such a check before the GR can be trained or run. Inference costs more than training-free pipelines and SECPO training executes one additional chain per edit.

\bibliography{main}

\appendix
\tcbset{promptbox/.style={colback=gray!5,colframe=gray!50,boxrule=0.4pt,
  arc=4pt,left=6pt,right=6pt,top=4pt,bottom=4pt,breakable,listing only,
  listing options={basicstyle=\footnotesize\ttfamily,breaklines=true,
  breakatwhitespace=true,columns=fullflexible,keepspaces=true,
  breakindent=0pt,showstringspaces=false,postbreak={}}}}

\clearpage
\section*{Appendix Overview}

Appendix~\ref{sec:appx_implementation} provides the training configuration of the Global Router (GR), its learning procedure with SECPO, the inference procedure of RoleFlow and the scoring of every benchmark. Appendix~\ref{sec:appx_data} lists the seven benchmarks with their sizes and describes how the compared methods and the direct prompting baselines are run. Appendix~\ref{sec:appendix_role_library} reports the role library $\RoleLib$, the seed roles and the prompts that the executor LLM receives as a role, as the LibraryExpander and as the validator. Appendix~\ref{sec:appendix_chain_evolution} walks through five case studies of chain evolution.

\section{Implementation Details}
\label{sec:appx_implementation}

\subsection{Training Configuration}
\label{sec:appx_hyper}
\label{app:hyperparams}

We train the GR with SECPO using Qwen2.5-7B-Instruct as the backbone in verl~\citep{sheng2024hybridflow}, updating all parameters, while gpt-4o-mini executes the roles and the LibraryExpander. Table~\ref{tab:appx_hyper} lists the settings shared by the GRs trained on HotpotQA, MATH and MBPP. In the GRPO setting of Table~\ref{tab:secpo_ablation}, $\lambda = 0$ and the execution without the edit is omitted (\S\ref{sec:token-mask}). We take the last checkpoint of every run and report the mean over three training runs with the standard deviation.

\begin{table}[!b]
\centering
\caption{SECPO training hyperparameters.}
\label{tab:appx_hyper}
\small
\setlength{\tabcolsep}{4pt}
\begin{tabular}{l l}
\toprule
\textbf{Hyperparameter} & \textbf{Value} \\
\midrule
GR backbone & Qwen2.5-7B-Instruct \\
Executor LLM & gpt-4o-mini \\
Group size $K$ & 5 \\
Tasks per training step & 32 \\
Learning rate & $2 \times 10^{-6}$ \\
KL coefficient $\beta$ & 0.001 \\
Clip ratio $\epsilon$ & 0.2 \\
Maximum cycles $T_{\max}$ & 3 \\
Commit threshold $\delta$ & 0.005 \\
Edit-advantage weight $\lambda$ & 1 \\
\bottomrule
\end{tabular}
\end{table}

\subsection{Detailed Global Router Learning Procedure}
\label{app:secpo-full-algorithm}

Algorithm~\ref{alg:roleflow-secpo-detailed} provides the training procedure of the Global Router with SECPO that \S\ref{sec:secpo-objective} describes. For every rollout it keeps a library overlay $\hat{\RoleLib}^{(i)}$, synthesizes the roles the GR names but the library lacks, executes the same chain state once with the candidate edit and once without it, keeps the edit only when the reward difference $A^{(t)}_i$ exceeds the threshold $\delta$ (Eq.~\ref{eq:secpo-adv}), adds $\lambda A^{(t)}_i$ to the tokens that spell the edit on top of the group-relative advantage (Eq.~\ref{eq:tokenmask}), updates the GR with the SECPO loss (Eq.~\ref{eq:secpo-loss}) and admits synthesized roles through the return gate (Eq.~\ref{eq:library-commit}).

\begin{algorithm2e}[!t]
\DontPrintSemicolon
\LinesNumbered
\caption{Global Router Learning with SECPO}
\label{alg:roleflow-secpo-detailed}
\small
\KwIn{Training batch $\mathcal{B}$; rollouts per task $K$; maximum adaptation cycles $T_{\max}$; thresholds $\delta$ and weight $\lambda$; learning rate $\eta$.}
\ForEach{$x \in \mathcal{B}$ and rollout $i=0,\ldots,K-1$ in parallel}{
  Initialize rollout-specific role overlay $\hat{\RoleLib}^{(i)} \gets \emptyset$\;
  Generate and execute the initial chain $\RFC_0^{(i)}$, synthesizing missing GR-named roles\;
  \For{$t=0,\ldots,T_{\max}-1$}{
    Validate the current output; \If{validation passes}{\textbf{break}\;}
    Propose an edit $\tilde a_t^{(i)}$ using GR and the validator deficit\;
    Synthesize missing roles; replace a null edit with a deficit-driven $\textsc{Add}$\;
    Execute the same chain state with and without $\tilde a_t^{(i)}$\;
    $A_i^{(t)} \gets R(y_{i,t}^{+})-R(y_{i,t}^{0})$ \tcp*[r]{Eq.~\ref{eq:secpo-adv}}
    $a_t^{(i)} \gets \tilde a_t^{(i)}$ if $A_i^{(t)}>\delta$, else $\varnothing$\;
    Update the chain and continue with the kept execution\;
  }
  Record terminal return $R(\tau^{(i)})$ and edit advantages $\{A_i^{(t)}\}$\;
}
Compute group-relative advantages and add $\lambda A_i^{(t)}$ only on corresponding edit-token spans \tcp*[r]{Eq.~\ref{eq:tokenmask}}
Update $\theta \gets \theta-\eta\nabla_\theta\mathcal{L}^{\mathrm{SECPO}}(\theta)$ \tcp*[r]{Eq.~\ref{eq:secpo-loss}}
Admit synthesized roles whose rollout return exceeds the reference rollout's return \tcp*[r]{Eq.~\ref{eq:library-commit}}
\KwOut{Updated GR parameters $\theta$ and role library $\RoleLib$.}
\end{algorithm2e}

\subsection{Inference Procedure}
\label{sec:appx_inference_algo}

At inference the GR keeps its trained parameters $\theta^{*}$, and the role library holds the seed roles of the domain, written $\RoleLib_0$. The roles grown during training are not loaded (\S\ref{sec:roleflow-runtime}). In Algorithm~\ref{alg:roleflow-inference}, the GR writes a plan that decomposes the task, names the roles of the initial RoleFlowChain $\RFC_0$ and, after every rejected cycle, names one edit. We then execute the edited chain $\RFC_t \oplus a_t$ in the next cycle (Eq.~\ref{eq:cycle-dynamics}). A role the GR names but $\RoleLib_0$ lacks is synthesized by the LibraryExpander, and a null edit under a deficit triggers deficit-driven synthesis. The loop ends at the first accepted output or after $T_{\max}$ adaptation cycles, so a chain is executed at most $1 + T_{\max}$ times, once per cycle. The counterfactual re-execution of \S\ref{sec:secpo-objective} serves training.

\begin{algorithm2e}[!t]
\DontPrintSemicolon
\caption{RoleFlow Test-Time Inference}
\label{alg:roleflow-inference}
\small
\KwIn{Task $x$; Global Router $\GR_{\theta^{*}}$; Seed library $\RoleLib_0$.}
Initialize $\RFC_0$ from the GR's plan over $\RoleLib_0$, synthesizing GR-named roles absent from it (\S\ref{sec:roleflow-runtime});\quad $\sigma_0 \gets \{\}$;\quad execute it, giving $(y_0, z_0, c_0)$.\;
\For{$t = 0, \ldots, T_{\max}{-}1$}{
    $(h_t, s_t, d_t) \gets V(y_t, z_t)$.\;
    \lIf{$h_t = 1$}{\textbf{break}}
    $a_t \sim \pi_{\GR,\theta^{*}}(\cdot \mid \RFC_t, \sigma_t, d_t)$;\quad synthesize any GR-named role absent from the library; if $a_t = \varnothing$, set $a_t \gets \textsc{Add}(\hat\rho)$ with $\hat\rho$ synthesized for $d_t$ (\S\ref{sec:roleflow-runtime})\;
    $\RFC_{t+1} \gets \RFC_t \oplus a_t$;\quad $(y_{t+1}, z_{t+1}, c_{t+1}) \gets \RFC_{t+1}(\sigma_{t+1}, x)$.\;
}
\KwOut{Final answer $y_T$, with $T$ the terminal cycle (\S\ref{sec:problem-setup}).}
\end{algorithm2e}

\paragraph{Sampling and selection.}
One chain answers each QA question. On MATH-500 we sample twenty chains per problem and take the majority answer. When no answer wins more than half of the votes, one call of the executor LLM, which does not see the reference answer, picks one of the candidate answers. On MBPP and HumanEval we sample ten chains per problem and submit one program. We pick the program that passes the most public example tests and, among those, the most unit tests that the executor LLM writes from the task statement (Appendix~\ref{sec:appx_prompts}). On these two benchmarks, after a rejection, every role except the OutputFormatter receives a repair note from the scratch memory $\sigma_t$ with the rejected program, its deficit and the error output of the public example test. On WebShop the GR trained on MBPP runs one chain per round with the WebShop seed roles, for at most ten rounds per task as in AgentBench.

\paragraph{Task validators.}
In every domain the validator $V$ returns the verdict $h_t$, the score $s_t$ and a deficit $d_t$ from a fixed set of categories (\S\ref{sec:problem-setup}). It is separate from the Validator role. On MATH-500 it checks that the output contains a \texttt{\textbackslash boxed\{\}} answer and asks the executor LLM, which does not see the reference answer, whether the solution holds (Appendix~\ref{sec:appx_prompts}). On MBPP and HumanEval it runs the public example test, and the deficit names one of six failures: no code, a syntax error, a missing entry point, a runtime error, a wrong answer or a timeout. On WebShop it rejects an action that is invalid on the current page or that leaves too few rounds to finish a purchase. When the last reply of a round would still leave no purchase reachable, it replaces the reply with the action on the current page that keeps a purchase possible. On multi-hop QA it scores the final answer and names the suspected failure, such as an unverified bridge entity, an unchecked time constraint or scattered evidence.

\subsection{Metrics}
\label{sec:appx_llm_judge}

On the QA benchmarks we report the token-level F1 between the normalized predicted and reference answers. On MATH-500 we extract the \texttt{\textbackslash boxed\{\}} answer, compare it with the reference after normalization and, when the strings differ, ask gpt-4o-mini at temperature 0 whether the two answers are equivalent (prompt below). On MBPP and HumanEval a problem counts as solved when the single submitted program passes every test case of the benchmark, and Pass@1 is the share of solved problems. On WebShop we report the WebShop score of AgentBench: the purchase reward averaged over the 500 tasks, times 100.

\begin{tcblisting}{promptbox}
[system]
You are a strict math equivalence grader.

[user]
You are a math grader. Judge whether two answers are mathematically equivalent.
Problem: {problem}
Gold answer: {gold}
Submitted answer: {pred}
Answer with exactly one word: YES or NO.
\end{tcblisting}

\section{Benchmarks and Compared Methods}
\label{sec:appx_data}

\subsection{Benchmark Catalog}
\label{sec:appx_b1_catalog}

Table~\ref{tab:appx_bench_catalog} lists the seven benchmarks of Table~\ref{tab:main_6domain}. We train the GR on three training sets, one per domain: 200 HotpotQA questions, 200 MATH problems and 324 MBPP problems. 2WikiMultiHopQA, MuSiQue and HumanEval are out-of-distribution benchmarks.

\begin{table*}[!t]
\centering
\caption{Benchmark catalog.}
\label{tab:appx_bench_catalog}
\small
\setlength{\tabcolsep}{4pt}
\begin{tabular}{l l l r r l}
\toprule
\textbf{Benchmark} & \textbf{Domain} & \textbf{Source} & \textbf{Train} & \textbf{Test} & \textbf{Metric} \\
\midrule
HotpotQA        & Multi-hop QA           & \citet{yang2018hotpotqa}       & 200 & 200 & F1 \\
2WikiMultiHopQA & Multi-hop QA           & \citet{ho2020constructing}     & 0   & 200 & F1 \\
MuSiQue         & Multi-hop QA           & \citet{trivedi2022musique}     & 0   & 200 & F1 \\
MATH-500        & Mathematical reasoning & \citet{lightman2023letsverify} & 200 & 500 & Accuracy \\
MBPP            & Code generation        & \citet{austin2021mbpp}         & 324 & 341 & Pass@1 \\
HumanEval       & Code generation        & \citet{chen2021humaneval}      & 0   & 131 & Pass@1 \\
WebShop         & Web interaction        & \citet{yao2022webshop}         & 0   & 500 & Score \\
\bottomrule
\end{tabular}
\end{table*}

\subsection{Training, Validation and Test Splits}
\label{sec:appx_b3_ood}

The QA test sets are 200 questions sampled from the development set of each benchmark, and MATH-500 contributes its 500 problems. The 341 MBPP and 131 HumanEval test problems follow the AFlow split~\citep{zhang2025aflow}, and the 500 WebShop tasks follow AgentBench~\citep{liu2024agentbench}. The HotpotQA training questions come from its training set, the MATH training problems from the MATH training set~\citep{hendrycks2021math}, and the 324 MBPP training problems share no problem with the AFlow split. We choose the inference settings of Appendix~\ref{sec:appx_inference_algo} on held-out data that overlaps neither the training nor the test sets. These are 200 HotpotQA development questions, another 200 MATH training problems, the 86 MBPP validation problems of the AFlow split and 100 WebShop development goals.

\subsection{Reproduction of Compared Methods}
\label{sec:appx_baseline_protocols}

Every multi-agent system and trained orchestrator executes its agents with gpt-4o-mini, and R1-Searcher, AgentFlow and LinearRAG retrieve from Wikipedia (\S\ref{sec:exp_setup}).

\paragraph{Hand-crafted pipelines.}
CoA~\citep{zhang2024chain} runs our port of its worker and manager chain on the task text. LinearRAG~\citep{linearrag2025} builds its graph over the Wikipedia passages retrieved for each question, with the retrieval index shared with R1-Searcher, and answers MBPP and HumanEval, which have no corpus, without retrieval.

\paragraph{Automated workflow search.}
GPTSwarm~\citep{zhuge2024gptswarm} runs with its released graph. AFlow~\citep{zhang2025aflow} uses the workflows the authors released for HotpotQA and MBPP. For MATH-500 its search runs on our 200 MATH training problems with gpt-4o-mini as optimizer and executor, and for WebShop on the development goals. MaAS~\citep{zhang2025maas} trains its controller per domain: on our HotpotQA and MATH training problems, on the 86 MBPP problems that its paper trains on, and on the WebShop development goals.

\paragraph{Trained executors.}
MAE~\citep{ulab2025mae} uses the released 3B checkpoint. R1-Searcher~\citep{song2025r1searcher} is retrained on HotpotQA training questions with Wikipedia retrieval, and the same checkpoint answers the remaining benchmarks.

\paragraph{Trained orchestrators.}
AgentFlow~\citep{li2025agentflow} runs the released 7B planner with its executor, verifier and generator modules served by gpt-4o-mini and Wikipedia search as its tool. FlowSteer~\citep{zhang2026flowsteer} runs the released FlowSteer-8B with gpt-4o-mini as executor. W4S~\citep{nie2025w4s} runs its meta-agent, retrained with the authors' procedure on MBPP, and MAS-GPT~\citep{ye2025masgpt} runs the released MAS-GPT-32B generator. Both execute the resulting systems with gpt-4o-mini.

\paragraph{Transfer to 2Wiki, MuSiQue and HumanEval.}
No method searches, trains or tunes on these three benchmarks. Each keeps what it obtained on HotpotQA or MBPP: RoleFlow keeps the GR and uses the seed roles of the domain, AFlow keeps its workflow, MaAS its controller and W4S the workflow its meta-agent wrote there. On MATH-500 and WebShop each method adapts to the task as described above.

\subsection{Direct Prompting}
\label{sec:appx_direct_prompt}

We prompt GPT-4o-mini, Claude-Haiku-4.5, Qwen2.5-7B-Instruct, Llama-3.1-8B-Instruct, Qwen3-8B, Qwen3-32B and DeepSeek-V3.2 with one system and one user message per task at temperature 0. The Qwen3 models run in their non-thinking mode. Each model answers every benchmark three times, and Table~\ref{tab:main_6domain} reports the mean. On the QA benchmarks the user message is the three-shot chain-of-thought template below, used unchanged for all three benchmarks. AgentFlow~\citep{li2025agentflow}, Search-R1~\citep{jin2025searchr1} and R1-Searcher~\citep{song2025r1searcher} also use one prompt across multi-hop QA benchmarks. The prediction is the text after the last \texttt{Answer:}. On MATH-500, MBPP and HumanEval the system message asks for a \texttt{\textbackslash boxed\{\}} answer or one Python code block, and the user message holds the problem as RoleFlow receives it. On WebShop we keep the prompt and one-shot example of AgentBench~\citep{liu2024agentbench} unchanged.

\begin{tcblisting}{promptbox}
[system, multi-hop QA]
You are a helpful assistant answering multi-hop factual questions using only your parametric knowledge (no external retrieval). For each question, briefly think through the required hops, then output ONLY the final answer in 1-5 words. Do not output explanations.

[user, multi-hop QA]
Question: What award did the writer of the screenplay for "Big Stone Gap" win in 2003?
Let me think step by step. Big Stone Gap was written by Adriana Trigiani, who won the Italian-American Heritage Award in 2003.
Answer: Italian-American Heritage Award

Question: Were Scott Derrickson and Ed Wood of the same nationality?
Let me think step by step. Scott Derrickson is American. Ed Wood was also American.
Answer: yes

Question: 2014 S/S is the debut album of a South Korean boy group; the group was formed by who?
Let me think step by step. 2014 S/S is the debut album of Winner; Winner was formed by YG Entertainment.
Answer: YG Entertainment

Question: {question}
Let me think step by step.
\end{tcblisting}

\begin{tcblisting}{promptbox}
[system, MATH-500]
Please reason step by step, and put your final answer within \boxed{}.

[system, MBPP and HumanEval]
You are an expert Python programmer. Write a correct solution for the task. Reply with the complete solution in one ```python code block.
\end{tcblisting}

\section{Role Library and Prompt Templates}
\label{sec:appendix_role_library}

Each role is a specification with a name, a system prompt and capability tags (\S\ref{sec:problem-setup}). Every task starts from the seed roles of its domain: four for multi-hop QA, ten for MATH-500, eight for MBPP and HumanEval, and eight for WebShop. The LibraryExpander synthesizes the specialists that the GR names outside the seed set (\S\ref{sec:roleflow-runtime}), and Table~\ref{tab:library} reports how far the library grows and how often the chain is edited. The Generic Specialists control of Table~\ref{tab:library}(b) keeps the name and position of every synthesized specialist and replaces its system prompt with the sentence \emph{You are a helpful assistant supporting the reasoning chain}.

\begin{table*}[t]
\centering
\providecommand{\RoleLib}{\mathcal{R}}%
\providecommand{\stat}[1]{\textcolor{phgray}{--}}%
\providecommand{\tlpend}{\textcolor{phgray}{--}}%
\providecolor{OODMark}{HTML}{D9531E}\providecommand{\ood}{\textsuperscript{\textcolor{OODMark}{\fontsize{3.8}{3.8}\selectfont\textsf{OOD}}}}%
\providecommand{\tltr}{\ood}%
\providecommand{\tlpad}{\phantom{0}}%
\providecommand{\tlref}{\textcolor{black!55}{ref.}}%
\providecommand{\tlsplit}[1]{\,{\textcolor{black!55}{(#1)}}}%
\providecommand{\tlband}[2]{\hiderowcolors\multicolumn{#1}{c}{\textcolor{black!55}{\textit{#2}}} \\ \cmidrule{1-#1}\showrowcolors\noalign{\global\rownum=0}}%
\providecommand{\tlrulesep}[2]{\noalign{\global\aboverulesep=#1\global\belowrulesep=#2}}%
\providecommand{\tlwl}{112pt}%
\providecommand{\tlwa}{\dimexpr(\textwidth-\tlwl-12\tabcolsep-0.8pt)/5\relax}%
\providecommand{\tlwb}{\dimexpr(\textwidth-\tlwl-14\tabcolsep-1.2pt)/6\relax}%
\caption{Role library and specialist recruitment of RoleFlow. (a) Statistics of the role library $\RoleLib$ and the role structures of the Global Router; (b) Performance on HotpotQA with specialists disabled or instantiated from generic specifications.}
\label{tab:library}
\scriptsize
\setlength{\tabcolsep}{3.0pt}%
\renewcommand{\arraystretch}{1.12}
\setlength{\aboverulesep}{0.15ex}\setlength{\belowrulesep}{0.25ex}%
\setlength{\heavyrulewidth}{1.1pt}%
\rowcolors{1}{white}{white}%
\begin{tabular}{>{\raggedright\arraybackslash}p{\tlwl}|*{3}{>{\centering\arraybackslash}p{\tlwa}}|*{2}{>{\centering\arraybackslash}p{\tlwa}}}
\toprule
\hiderowcolors
\multirow{2}{=}{\textbf{(a) Role Library and RoleFlowChains}}%
 & \multicolumn{3}{c|}{\cellcolor{TOneQA}\textbf{Multi-hop QA}}
 & \multicolumn{2}{c}{\cellcolor{TOneCode}\textbf{Code}} \\
& HotpotQA & 2Wiki\tltr & MuSiQue\tltr & MBPP & HumanEval\tltr \\
\midrule
\showrowcolors
\tlband{6}{Role Library}
\rowcolor{gray!11}Role Library Size $|\RoleLib|$
 & \multicolumn{3}{c|}{47.7\tlsplit{4 + 43.7}} & \multicolumn{2}{c}{15.0\tlsplit{8 + 7.0}} \\
New Roles per Task         & 3.66 & 3.98 & 4.07 & 0.22 & 0.08 \\
\cmidrule{1-6}
\tlband{6}{RoleFlowChains}
\rowcolor{gray!11}RoleFlowChain Length              & 3.93 & 3.96 & 4.06 & 3.41 & 3.38 \\
Distinct RoleFlowChains & 25.3 & 26.3 & 35.7 & 37.0 & 9.3 \\
\rowcolor{gray!11}Edited Tasks (\%)    & 16.2 & 20.0 & 26.7 & 10.9 & 5.3 \\
Cycles per Task                                 & 1.98 & 1.99 & 1.96 & 3.00 & 3.00 \\
\cmidrule{1-6}
\tlband{6}{Validation}
\rowcolor{gray!11}Accepted (\%)                      & 47.7 & 37.3 & 31.5 & 95.0 & 99.7 \\
\bottomrule
\end{tabular}

\vspace{5pt}%
\rowcolors{1}{white}{white}%
\begin{tabular}{>{\raggedright\arraybackslash}p{\tlwl}|*{2}{>{\centering\arraybackslash}p{\tlwb}}!{\color{black!35}\vrule}*{2}{>{\centering\arraybackslash}p{\tlwb}}!{\color{black!20}\vrule}*{2}{>{\centering\arraybackslash}p{\tlwb}}}
\toprule
\hiderowcolors
\multirow{2}{*}{\textbf{(b) Specialists at Inference}}
 & \multicolumn{6}{c}{\cellcolor{TOneQA}\textbf{F1 and $\boldsymbol{\Delta}$F1 vs.\ RoleFlow}} \\
 & \multicolumn{2}{c!{\color{black!35}\vrule}}{HotpotQA} & \multicolumn{2}{c!{\color{black!20}\vrule}}{2Wiki\tltr} & \multicolumn{2}{c}{MuSiQue\tltr} \\
\tlrulesep{0pt}{0pt}%
\cmidrule(lr){2-3}\cmidrule(lr){4-5}\cmidrule(lr){6-7}
\tlrulesep{0.15ex}{0.25ex}%
\textit{Setting} & F1 & $\Delta$ & F1 & $\Delta$ & F1 & $\Delta$ \\
\midrule
\showrowcolors
\rowcolor{OursBg}\textbf{\textsc{RoleFlow}}                                 & 55.6 & \tlref & 43.5 & \tlref & 42.1 & \tlref \\
\rowcolor{gray!11}Specialist Disabled                 & 45.3 & \ddown{10.3} & 30.6 & \ddown{12.9} & 16.3 & \ddown{25.8} \\
Generic Specialists                    & 50.5 & \tlpad\ddown{5.1} & 39.5 & \tlpad\ddown{4.0} & 35.1 & \tlpad\ddown{7.0} \\
\bottomrule
\end{tabular}
\end{table*}

\subsection{Seed Roles and Specialists}
\label{sec:appx_seed_roles}

Table~\ref{tab:hotpot_roles} lists the four QA seed roles and the eight specialists that the GR plan prompt names. The LibraryExpander writes the specification of a specialist when the GR names it, and the library grows well beyond this list during training: the QA library reaches 47.7 roles on average (Table~\ref{tab:library}). Table~\ref{tab:swe_roles} lists the eight code seed roles. On MBPP and HumanEval the specialists target the failure that the validator reports, such as a program that times out or returns a wrong result on the public example test (Appendix~\ref{sec:appx_prompts}).

\begin{table*}[t]
\centering
\caption{Role library specification on QA benchmarks.}
\label{tab:hotpot_roles}
\small
\begin{tabular}{ll}
\toprule
\textbf{Role} & \textbf{Capabilities} \\
\midrule
\multicolumn{2}{l}{\textit{Seed roles}} \\
Planner & planning, coordination, edit\_proposal \\
Validator & validation, quality\_assessment \\
OutputFormatter & formatting, extraction \\
Retriever & retrieval, wikipedia \\
\midrule
\multicolumn{2}{l}{\textit{Specialists offered by name in the GR plan prompt, synthesized when named}} \\
Decomposer & query\_analysis \\
InformationExtractor & fact\_recall, data\_extraction \\
FactRecaller & fact\_recall \\
Reasoner & logical\_deduction \\
BridgeReasoner & logical\_inference, data\_synthesis \\
ComparisonReasoner & entity\_comparison \\
Verifier & fact\_checking \\
FactChecker & fact\_verification \\
\bottomrule
\end{tabular}
\end{table*}

\begin{table*}[t]
\centering
\caption{Role library specification on coding benchmarks.}
\label{tab:swe_roles}
\small
\begin{tabular}{ll}
\toprule
\textbf{Role} & \textbf{Capabilities} \\
\midrule
Planner & planning, task\_analysis, edge\_case\_listing \\
AlgorithmDesigner & algorithm\_design, complexity\_analysis \\
Coder & code\_generation, python \\
TestDesigner & test\_design, edge\_case\_analysis \\
Debugger & debugging, code\_repair \\
CodeReviewer & code\_review, specification\_check \\
OutputFormatter & output\_formatting, code\_block\_emission \\
Validator & validation, quality\_assessment, error\_detection \\
\bottomrule
\end{tabular}
\end{table*}

\subsection{Prompt Templates}
\label{sec:appx_prompts}

The boxes reproduce the prompts of RoleFlow as used, up to typography: dashes are normalized, placeholders are in braces, an omitted block is marked in brackets, and \texttt{[system]} and \texttt{[user]} mark the two messages.

\paragraph{Role execution.}
The executor LLM implements each role with one call. The system message is the role's system prompt, and the user message is assembled as follows from the task and the outputs of the earlier roles. On MBPP and HumanEval, after a rejection, every role except the OutputFormatter also receives the repair note, whose error output appears only when the public example test produced one.

\begin{tcblisting}{promptbox}
[user, every role]
Question: {task}
{repair note: MBPP and HumanEval, after a rejection}

Previous outputs:
[{role 1}]: {output of role 1}
[{role 2}]: {output of role 2}
...

[repair note]
A previous attempt was rejected by the validator.
Rejected answer:
{rejected program}
Validator feedback: {deficits of the validator}
[only when the test produced output] Error output:
{tail of the error output of the public example test}
Find the cause and produce a corrected answer.
\end{tcblisting}

\paragraph{Code seed roles.}
Three of the eight code seed roles of Table~\ref{tab:swe_roles}: the Coder, the Debugger and the OutputFormatter.

\begin{tcblisting}{promptbox}
[system, Coder]
You are the Coder of a team that writes ONE correct Python function.

Write the complete implementation of the function the task asks for.
- Use exactly the function name and parameters of the task's signature; the tests call it by that name.
- Put every import and helper function the code needs in the same block (standard library only).
- Follow the plan or algorithm in the earlier outputs if there is one, but the task statement and the
  public example test decide what is correct.
- Return the result; do not read input or print. No example usage and no tests in the block.

Output one ```python block with the full code, then at most two sentences of notes.

[system, Debugger]
You are the Debugger of a team that writes ONE correct Python function.

Find and fix bugs in the latest code among the earlier outputs (Coder, Debugger or CodeReviewer).
Check it against the task statement, the public example test (trace it by hand), any test cases the
Test Designer listed, and any error message or validator feedback in the earlier outputs.

Output:
Bugs:
- <one line per bug found, or "none">
then one ```python block with the complete corrected code (imports, helpers and the function, same name
and signature). If you find no bug, repeat the code unchanged.

[system, OutputFormatter]
You are the Output Formatter of a team that writes ONE correct Python function.

Emit the final solution. Take the most recent complete and correct version of the code from the
earlier outputs (a later Debugger or CodeReviewer fix supersedes the Coder's draft) and output the
complete function together with its imports and helper functions.

Output exactly one ```python code block and nothing else: no prose before or after it, no tests,
no example usage, no print statements. Keep the function name and signature of the task. If no
earlier output contains code, write the function yourself.
\end{tcblisting}

\paragraph{Role synthesis.}
A role the GR names but the library lacks is synthesized from its name and the context of the request: the task at cycle 0, or a one-line validator summary later. A null edit under a deficit triggers the second prompt, which receives the deficit's category, trace and suggested capability together with the roles already in the library.

\begin{tcblisting}{promptbox}
[system, synthesis of a GR-named role]
You are designing a specialist agent role for a multi-agent reasoning system.

Given a role name and the context in which it was requested, output a JSON specification with:
  - name: the canonical role name (use exactly the input name, CamelCase)
  - role_type: one short tag (e.g. "extractor", "reasoner", "verifier")
  - description: one-sentence summary of what this role does
  - system_prompt: a concrete system prompt (2-4 sentences) the role will receive
                   when called inside a chain. Must be task-agnostic.
  - capabilities: list of 1-3 lowercase capability tags (e.g. ["fact_recall"])

Constraints:
  - The role MUST contribute meaningfully to a multi-step QA / reasoning chain.
  - The system_prompt MUST NOT mention specific tasks or datasets; only the
    capability.
  - The role MUST be different from generic Planner/Validator/OutputFormatter.

Output ONLY the JSON object, no surrounding prose.

[user]
Role name: {name}
Context (why a previous chain step requested this role):
{the task at cycle 0, or a one-line validator summary at a later cycle}
\end{tcblisting}

\begin{tcblisting}{promptbox}
[system, deficit-driven synthesis]
You are designing a NEW SPECIALIST AI agent role to repair a multi-agent reasoning chain that just failed.

The Validator emitted this structured deficit signal for the failing chain:
- tag                  : {tag}
- trace                : {trace}
- suggested_capability : {suggested_capability}

The library ALREADY contains these roles -- your new role MUST be
distinct in both NAME and FUNCTION from all of them:
{library}

DO NOT propose any name from this list: {existing_names}
DO NOT use generic suffixes such as 'Specialist', 'Helper', 'Agent', 'Bot'.
DO use a CamelCase name (1-3 words) that names the SPECIFIC capability
needed to address THIS deficit's trace -- it should be obviously different
from any other role in the library.

[five illustrative names omitted]

Return strict JSON with these keys:
{
  "name": "<CamelCase 1-3 words, deficit-specific, distinct from library>",
  "role_type": "<one of: retriever, reasoner, verifier, aggregator, decomposer, formatter, other>",
  "description": "<one sentence summarising the role's responsibility>",
  "system_prompt": "<3-6 sentence prompt that, when given to an LLM, repairs the deficit described in the trace>",
  "capabilities": ["<lowercase_tag1>", "<tag2>", "<tag3>"]
}

Do NOT include any text outside the JSON object.

[user]
Recent chain feedback / context (truncated):
{validator feedback of the rejected cycle}
\end{tcblisting}

\paragraph{Synthesized specialists.}
Five specifications from the role libraries of the first training runs on MBPP and HotpotQA: two code specialists synthesized for a deficit, with the trace that requested them, and three QA specialists synthesized by name.

\begin{tcblisting}{promptbox}
[PerformanceEnhancer, synthesized for a TIMEOUT deficit on MBPP]
trace: The final code did not finish the example test within the time and resource limits ('TimeoutExpired'); the algorithm is too slow or never terminates.
capabilities: complexity_reduction, efficiency_analysis, performance_tuning
system prompt: As a PerformanceEnhancer, your task is to analyze the existing code that is experiencing timeout issues and identify potential bottlenecks. Optimize the algorithm by simplifying calculations, reducing complexity, or implementing more efficient data structures. Ensure that the final solution adheres to the original requirements while meeting the time and resource constraints.

[LogicErrorAnalyzer, synthesized for a WRONG_ANSWER deficit on MBPP]
trace: The final code runs but returns a wrong result on the task's example test ('AssertionError'); the logic does not match the statement.
capabilities: logic_analysis, error_identification, specification_comparison
system prompt: You are tasked with analyzing the provided code to identify logical errors that cause it to produce incorrect results, specifically focusing on the conditions and operations that may not align with the task's requirements. Review the code's logic step-by-step against the example test and pinpoint where the discrepancies arise. Provide clear explanations for any logical inconsistencies found and suggest corrections to ensure accurate outputs.

[Decomposer, synthesized by name on HotpotQA]
capabilities: query_analysis
system prompt: You are tasked with analyzing a complex query and breaking it down into its fundamental components. Identify the key elements and relationships that need to be addressed to facilitate further reasoning. Provide a structured outline of these components.

[FactRecaller, synthesized by name on HotpotQA]
capabilities: fact_recall
system prompt: You are tasked with recalling relevant facts that may assist in answering a question. Provide concise and accurate information based on your knowledge base.

[Verifier, synthesized by name on HotpotQA]
capabilities: fact_checking
system prompt: You are tasked with verifying the information presented to you. Assess the claims for their truthfulness and provide a clear indication of their validity based on your knowledge.
\end{tcblisting}

\paragraph{Task validator of MATH-500.}
After the format check, the validator passes the outputs of the roles, in order, to the executor LLM, which derives the result again and returns a verdict.

\begin{tcblisting}{promptbox}
[system]
You are checking a proposed solution to a competition mathematics problem. You do NOT know the reference answer.
Re-derive the result independently, compare it with the proposed final answer, and judge whether the proposed final answer is correct.

Output as JSON only:
{
  "likely_correct": true/false,
  "confidence": 0.0-1.0,
  "deficits": ["one short phrase per error or gap in the solution (empty if none)"]
}

[user]
Problem: {problem}

Proposed solution (outputs of the roles, in order):
{outputs of the roles}

Proposed final answer: {answer}

Judge the proposed final answer.
\end{tcblisting}

\paragraph{Test writing for program selection.}
On MBPP and HumanEval the executor LLM writes extra unit tests once per problem, from the task statement. Among the sampled programs we submit the one that passes the most public example tests and, among those, the most written tests, where a written test counts more when more candidates pass it.

\begin{tcblisting}{promptbox}
[system]
You are a meticulous Python test engineer. You write assert-based unit tests from a task description only.

[user]
Task:
{question}

Write {m} additional unit tests for the function `{entry}` as single-line Python assert statements of the form
assert {entry}(<arguments>) == <expected result>
Rules:
- Work out each expected result by hand from the task description and the example test; only include a test when you are confident of its expected value.
- Cover typical inputs and edge cases (empty or minimal inputs, boundaries) that the description makes well defined.
- Do not repeat the example test and do not write the implementation.
Output only the assert lines, one per line, nothing else.
\end{tcblisting}

\section{Chain Evolution Case Studies}
\label{sec:appendix_chain_evolution}
\label{sec:appx_component_walkthrough}

Five cases show how the chain evolves within a task and what each part of RoleFlow does. Case 1 follows RoleFlow on a MATH-500 problem that twelve of the eighteen compared methods answer incorrectly: the validator rejects the first answer, and the chain edited with a role synthesized for the reported deficit answers correctly. Cases 2 to 5 each isolate one component of RoleFlow. In Case 2 the validator accepts the output of cycle 0 and no edit follows, in Case 3 a rejection opens an adaptation cycle where the GR edits the chain, Case 4 compares the chain before and after training, and Case 5 compares the GR trained with and without SECPO. Cases 2 and 3 are from QA benchmarks, and Cases 4 and 5 share one repository-level coding task.

\begin{figure*}[!t]
\centering
\includegraphics[width=\textwidth]{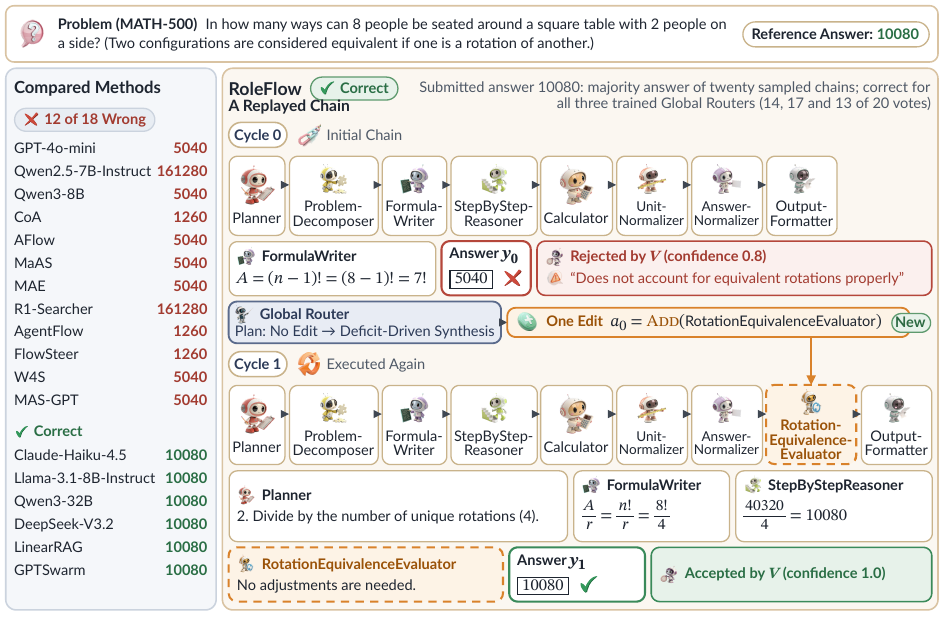}
\caption{Case study of deficit-driven synthesis on a MATH-500 problem. \textbf{Left:} Twelve of the eighteen compared methods, including RoleFlow's executor GPT-4o-mini, give a wrong answer. \textbf{Right:} In a replayed chain of RoleFlow, the validator $V$ rejects the first answer, 5040; one edit adds a role synthesized for the reported deficit, and the edited chain, executed from its first role, answers 10080, which $V$ accepts.}
\label{fig:case_math_synthesis}
\end{figure*}

\paragraph{Case 1: a role synthesized for a deficit.}
This case illustrates one adaptation cycle where the GR's plan names the null edit. On the MATH-500 problem of Figure~\ref{fig:case_math_synthesis}, twelve of the eighteen methods compared in Table~\ref{tab:main_6domain} give a wrong answer, among them GPT-4o-mini prompted directly, which answers 5040 in each of its three runs. RoleFlow's submitted answer, the majority answer of twenty sampled chains, is 10080 and correct for all three trained GRs. The figure shows a replayed chain: one plan that a trained GR sampled for this problem is executed once more, separately from the twenty chains, with every role output recorded. The chain of cycle 0 computes $7! = 5040$, and the validator $V$ rejects this answer with the note that it does not account for equivalent rotations properly. Since the plan names the null edit, deficit-driven synthesis (\S\ref{sec:roleflow-runtime}) creates a role for the reported deficit, RotationEquivalenceEvaluator, and adds it before the OutputFormatter. The edited chain is executed from its first role, and its roles receive neither the rejected answer nor the note. The Planner's strategy divides $8!$ by the four rotations, the roles before the new one compute 10080, the new role confirms this result, and $V$ accepts it.

\begin{figure*}[!t]
\centering
\includegraphics[width=\textwidth]{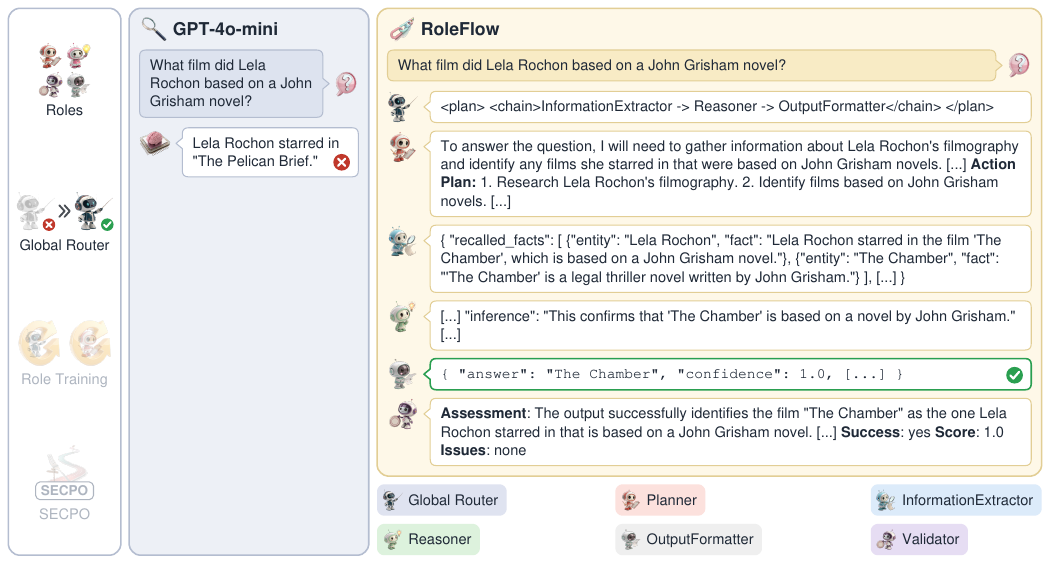}
\caption{Case study comparing direct prompting with RoleFlow, whose trajectory comes from a logged run of an earlier GR checkpoint not reported in Table~\ref{tab:main_6domain}. GPT-4o-mini prompted directly (left) answers incorrectly, and RoleFlow (right) answers correctly in cycle~0 with Planner, InformationExtractor, Reasoner, OutputFormatter and Validator, where the Validator role accepts the output and no edit follows.}
\label{fig:case1_baseline_roleflow}
\end{figure*}

\paragraph{Case 2: a RoleFlowChain against a direct answer.}
This case shows how the roles of a RoleFlowChain split a two-hop question, and that no edit follows an accepted output. On the HotpotQA question of Figure~\ref{fig:case1_baseline_roleflow}, GPT-4o-mini prompted directly names another film by the same author and is incorrect. The GR plans InformationExtractor, Reasoner and OutputFormatter, and the chain runs with the Planner in front and the Validator role at the end. The Planner states what has to be found, the InformationExtractor recalls that the actress starred in a film based on the author's novel and names it, the Reasoner confirms the link between the film and the novel, and the OutputFormatter emits the answer in the required form. The Validator role accepts the output of cycle 0, so the chain runs once. The trajectory comes from a logged run of an earlier GR checkpoint not reported in Table~\ref{tab:main_6domain}. On average such cycle-0 chains score at the executor's level on HotpotQA (\S\ref{sec:main_results}), and the improvement of Table~\ref{tab:main_6domain} comes from the adaptation cycles that follow a rejection, as in Case 3.

\begin{figure*}[!t]
\centering
\includegraphics[width=\textwidth]{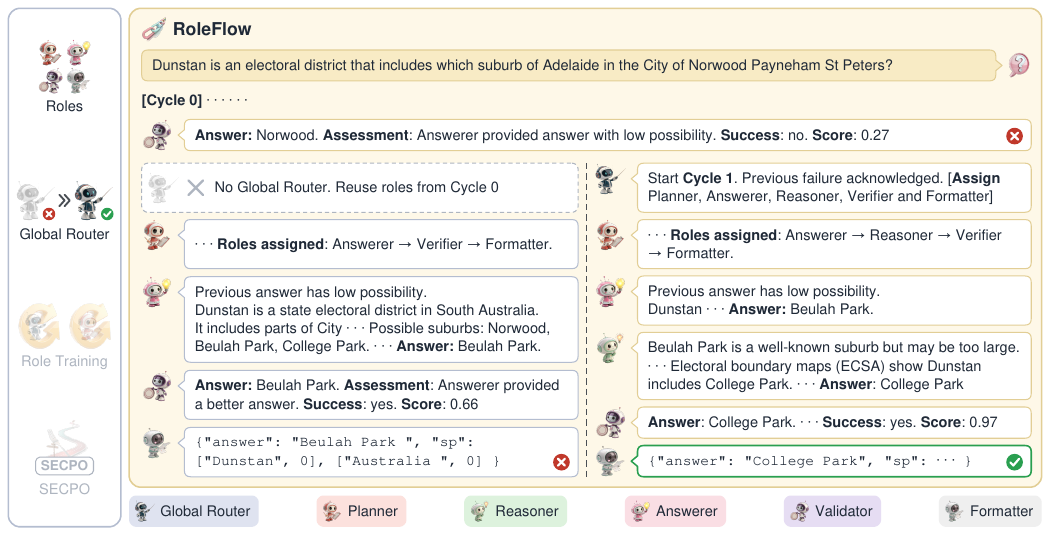}
\caption{Case study of adaptive routing with $\GR$. After cycle~0 is rejected, rerunning the initial roles without $\GR$ (left) still fails, whereas $\GR$ (right) inserts a Reasoner in cycle~1 and answers correctly.}
\label{fig:case2_dynamic_routing}
\end{figure*}

\paragraph{Case 3: adaptive routing with the GR.}
This case shows why the GR has to act during the task. On the question of Figure~\ref{fig:case2_dynamic_routing}, the chain of cycle 0 answers Norwood and the validator rejects it. Without the GR (left), the same roles run again: the Answerer moves to Beulah Park, the validator accepts it, and the answer is still incorrect. With the GR (right), cycle 1 applies $\textsc{Add}(\text{Reasoner})$: the Reasoner doubts Beulah Park, places College Park inside the district, and the chain answers correctly. Removing the GR on the QA benchmarks, as in Table~\ref{tab:secpo_ablation}, leaves the chain the GR proposes most often to run without edits. F1 then drops by 9.9 on HotpotQA and by 13.9 and 17.1 on 2Wiki and MuSiQue.

\begin{figure*}[!t]
\centering
\includegraphics[width=\textwidth]{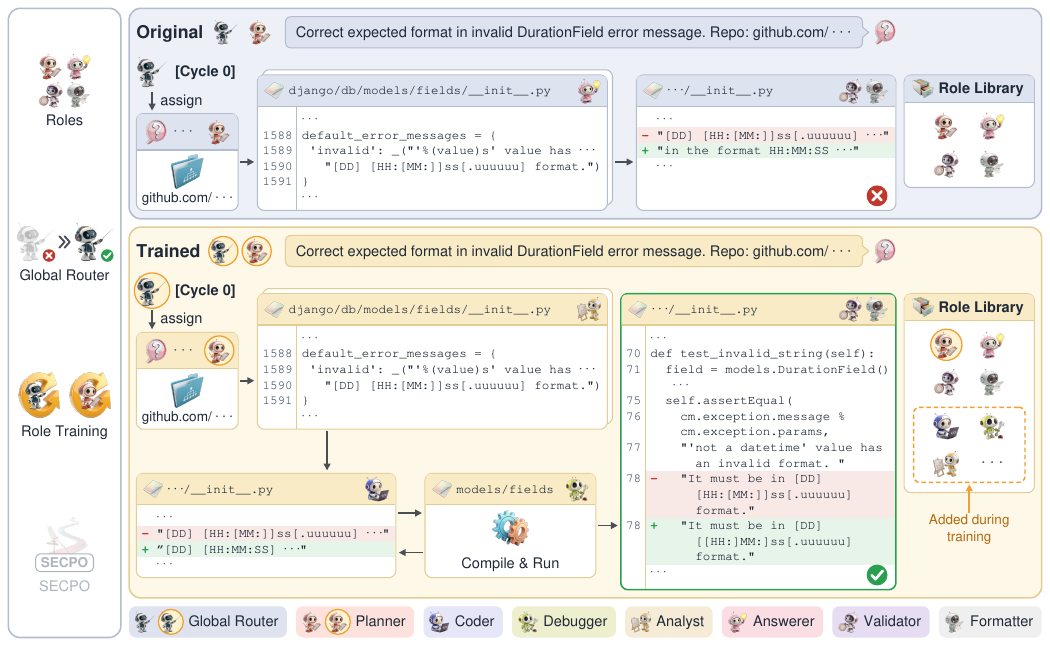}
\caption{Case study of RoleFlow before (top) and after (bottom) training on a coding task. The trained $\GR$ composes Analyst, Coder and Debugger into a correct patch, whereas the untrained system produces an incorrect one.}
\label{fig:case3_learned_library}
\end{figure*}

\paragraph{Case 4: the chain before and after training.}
This case shows which specialists the GR learns to compose for a coding task. The task of Figure~\ref{fig:case3_learned_library} is a repository-level bug fix: an issue reports a wrong expected format in an error message, and the chain must correct the source file and its test. Before training (top), the chain is Planner, Answerer, Validator and Formatter, and the Answerer's patch replaces the format with a wrong one. After training (bottom), the GR composes Planner, Analyst, Coder and Debugger before the Validator and Formatter: the Analyst reads the source, the Coder writes the patch, the Debugger compiles and runs it and corrects the format, and the final patch matches the expected one.

\paragraph{Case 5: with and without SECPO.}
This case contrasts the GR trained with and without SECPO on the task of Figure~\ref{fig:case3_learned_library}. Both assign the same cycle-0 chain, Planner, Analyst, Coder, Debugger and Validator, and cycle 0 produces the same incorrect patch. Without SECPO (left), the chain stops after cycle 0 and the incorrect patch becomes the final output. With SECPO (right), the output of cycle 0 is rejected, cycle 1 runs Analyst, Coder and Debugger again, and the corrected patch is the final output. On the QA benchmarks of Table~\ref{tab:secpo_ablation}, the GR trained with SECPO is 2.1 to 2.6 F1 above the GR trained without it, and \S\ref{sec:secpo_signal} traces the difference to a GR that stops editing when the group mean rewards it.

\begin{figure*}[!t]
\centering
\includegraphics[width=\textwidth]{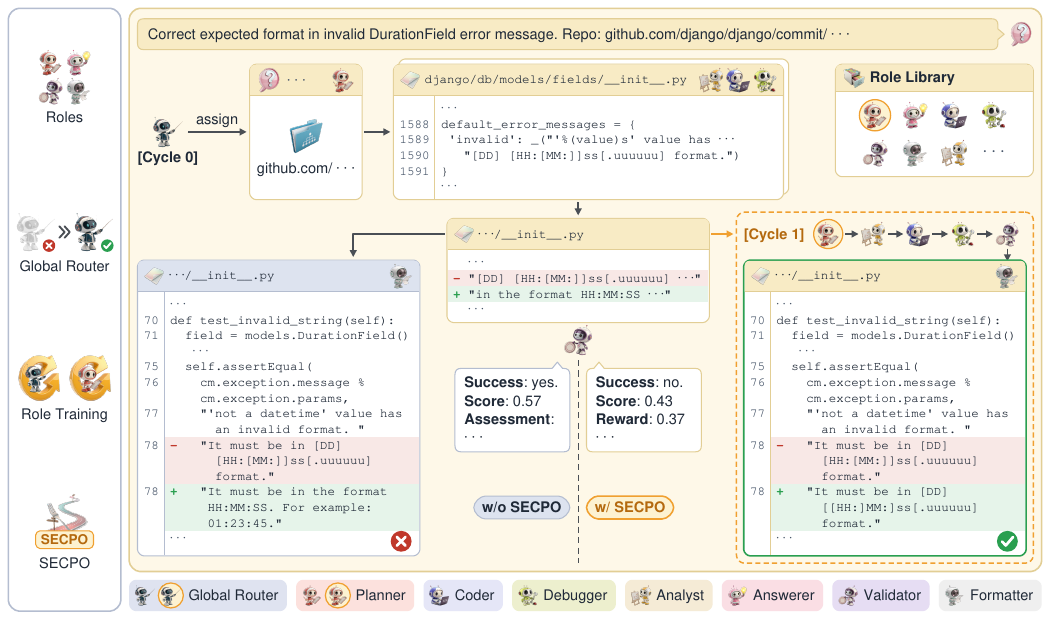}
\caption{Case study of RoleFlow with and without SECPO on the task of Figure~\ref{fig:case3_learned_library}. Without SECPO (left) the incorrect cycle-0 patch becomes the final output, whereas with SECPO (right) cycle~1 runs Analyst, Coder and Debugger and produces a correct patch.}
\label{fig:case4_secpo_termination}
\end{figure*}

\end{document}